\pdfoutput=1
\documentclass[letterpaper]{article} 
\usepackage{aaai2027}

\usepackage[hyphens]{url}  
\usepackage{graphicx} 
\usepackage{natbib}  
\usepackage{caption} 
\usepackage{algorithm}

\nocopyright
\usepackage{newfloat}
\usepackage{listings}
\DeclareCaptionStyle{ruled}{labelfont=normalfont,labelsep=colon,strut=off} 
\floatstyle{ruled}
\newfloat{listing}{tb}{lst}{}
\floatname{listing}{Listing}

\usepackage{booktabs}

\title{OrchBench: Evaluating Multi-Agent Orchestration Plans in Isolation via \\ Deterministic Simulation}

\author{
    Zhenzhen Ren\textsuperscript{\rm 1, \rm 2},
    Jiyan He\textsuperscript{\rm 2},
    Xinpeng Zhang\textsuperscript{\rm 1},
    Zhenxing Qian\textsuperscript{\rm 1},\\
    Ke Han\textsuperscript{\rm 3},
    Shuxin Zheng\textsuperscript{\rm 2},
    GuoBiao Li\textsuperscript{\rm 1},
    Xiaoqing Zhang\textsuperscript{\rm 2}\corresponding
}

\affiliations{
    \textsuperscript{\rm 1}Fudan University\\
    \textsuperscript{\rm 2}Zhongguancun Academy\\
    \textsuperscript{\rm 3}Queen Mary University of London\\
}

\newcommand{\bench}{OrchBench}

\usepackage{url}
\usepackage{booktabs}
\usepackage{multirow}
\usepackage{algorithm}
\usepackage{algpseudocode}
\usepackage{amsmath}
\usepackage{amssymb}
\usepackage{graphicx}
\usepackage{xcolor}
\usepackage{tikz}
\usetikzlibrary{arrows.meta, positioning}

\usepackage{tabularx}

\usepackage{tikz}
\usepackage{xcolor}
\usepackage{subcaption}

\usepackage{xcolor}
\usepackage{listings}
\usepackage{inconsolata}

\definecolor{codebg}{HTML}{F6F8FA}
\definecolor{codeborder}{HTML}{D0D7DE}
\definecolor{codestring}{HTML}{1F4E79}

\lstdefinestyle{workflowjson}{
  basicstyle=\ttfamily\small,
  stringstyle=\color{codestring},
  morestring=[b]",
  backgroundcolor=\color{codebg},
  frame=single,
  rulecolor=\color{codeborder},
  framesep=5pt,
  showstringspaces=false,
  columns=fullflexible,
  keepspaces=true,
  breaklines=true,
  breakatwhitespace=false,
  tabsize=2,
  aboveskip=3pt,
  belowskip=3pt
}

\usepackage{listings}
\usepackage{placeins}
\usepackage[most]{tcolorbox}
\tcbuselibrary{listings,breakable}

\newtcblisting{promptbox}[1]{
  breakable,
  listing only,
  colback=gray!3,
  colframe=black!35,
  arc=1mm,
  boxrule=0.4pt,
  left=1mm,
  right=1mm,
  top=1mm,
  bottom=1mm,
  title={#1},
  fonttitle=\bfseries,
  listing options={
    basicstyle=\ttfamily\footnotesize,
    breaklines=true,
    columns=fullflexible,
    keepspaces=true
  }
}

\begin{document}
\maketitle

\begin{abstract}
Complex tasks often decompose into parallelizable yet interdependent subtasks, making orchestration critical to the performance of multi-agent systems (MAS). 
Existing evaluations typically rely on end-to-end execution, which conflates orchestration-plan quality with worker capabilities, tool reliability, and environmental noise.
Moreover, the time and token costs of real execution grow rapidly with workflow scale, making systematic evaluation expensive.
We present OrchBench, a simulation-based benchmark for evaluating multi-agent orchestration plans in isolation. Starting from real-world tasks, OrchBench constructs directed acyclic graphs (DAGs) that encode task dependencies, with controlled sizes and degrees of parallelism. 
Given a DAG, a per-agent context limit, and an agent budget, the evaluated planner assigns subtasks to agents and
specifies cross-agent information transfers and their retention ratios. 
A deterministic simulator evaluates the resulting plan without invoking worker agents and returns interpretable measures of result quality, makespan, and token cost.
The simulated scores produced by OrchBench correlate strongly with quality scores from Claude Code executions, achieving a Pearson correlation of \(r=0.816\), while requiring only \(1.3\%\) of the tokens and \(10.3\%\) of the wall-clock time. 
Across diverse planners and workflow scales, we find
that preserving task-critical information is more important than simply increasing the number of agents, and the benefits of parallelism diminish as coordination failures accumulate. 
These results establish OrchBench as an efficient and interpretable benchmark for comparing and diagnosing multi-agent orchestration plans.
\end{abstract}

\begin{figure}[t]
\centering
\includegraphics[width=\columnwidth]{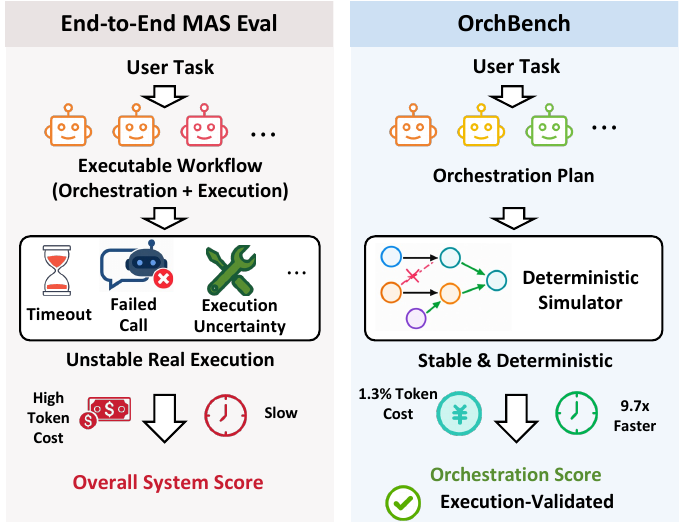}
\caption{Comparison between conventional end-to-end MAS evaluation and \bench{}. Instead of executing the complete multi-agent system, \bench{} evaluates orchestration plans with a deterministic simulator, isolating orchestration quality from execution-related factors while enabling fast, low-cost, and reproducible evaluation.}
\label{fig:why-simulation}
\end{figure}

\section{Introduction}

Multi-agent systems (MAS) have emerged as a powerful paradigm for solving complex tasks by coordinating multiple specialized agents \citep{guo2024large,hong2024metagpt,wang2402tdag,fourney2024magentic}.
As these systems become increasingly sophisticated, their performance depends not only on the capability of individual agents, but also on the orchestration strategy, i.e., how a workflow is organized across multiple agents.
Even with the same task and the same set of agents, different orchestration strategies can lead to substantially different outcomes \citep{yu2025dyntaskmas,kim2023llm,zhuge2024gptswarm}.
This growing importance raises a fundamental question: how can orchestration itself be evaluated?

However, evaluating orchestration itself remains challenging.
Existing benchmarks primarily evaluate the end-to-end task performance of MAS in realistic environments \citep{liu2023agentbench,mialon2023gaia,zhou2023webarena,jimenez2023swebench,xie2024osworld,koh2024visualwebarena,drouin2024workarena,yoran2024assistantbench,xu2024theagentcompany,deng2023mind2web}.
The resulting performance conflates orchestration quality with execution-related factors 
\citep{ma2024agentboard,ke2026mas,tsang2026reward}, making it difficult to attribute success or failure to orchestration itself, as illustrated in Figure~\ref{fig:why-simulation}.
Moreover, end-to-end evaluation is computationally expensive, making systematic comparison across orchestration strategies increasingly impractical as workflow scale grows.
Consequently, a practical evaluation framework that isolates orchestration from worker execution remains absent.

To address this issue, we introduce \bench{}, a \emph{simulation-based} benchmark for evaluating multi-agent orchestration independently of worker execution.
Our key insight is that orchestration can be evaluated without executing the complete multi-agent system.
Instead of evaluating end-to-end task performance, we require the model to generate only an orchestration plan.
Worker execution is then replaced by a deterministic simulator, which removes the influence of worker reasoning, tool execution, and environmental noise from the evaluation while enabling different orchestration strategies to be compared under identical conditions.
To enable systematic analysis of orchestration across different workflow scales, \bench{} contains semantically grounded dependency workflows ranging from 10 to 1,000 subtasks.
Beyond producing evaluation scores, the simulator explicitly identifies coordination failures, providing interpretable diagnoses for improving orchestration strategies.

Using \bench{}, we conduct the first systematic study of orchestration strategies produced by language models.
Our experiments uncover a previously hidden coordination bottleneck: as workflow scale increases, orchestration quality becomes increasingly constrained by information preservation rather than agent count and the benefits of parallelism decrease as coordination failures accumulate.
These failures are largely obscured by conventional end-to-end evaluation but become directly observable through our simulator.

We further validate that the simulator faithfully reflects real-world orchestration quality.
Across Claude Code executions, simulator scores achieve a strong correlation with real execution quality (Pearson \(r=0.816\)), while requiring only \(1.3\%\) of the tokens and \(10.3\%\) of the wall-clock time.
Beyond evaluation, simulator-guided diagnosis consistently improves real multi-agent execution by identifying missing information transfers.
Our contributions are threefold:
\begin{itemize}

    \item We identify the need to evaluate multi-agent orchestration independently of worker execution and formulate it as a standalone benchmark problem.

    \item We introduce \bench{}, a scalable benchmark that combines semantically grounded task workflows with a deterministic simulator for efficient and reproducible orchestration evaluation.


    \item We conduct a systematic study of orchestration strategies across workflows containing up to 1,000 subtasks, revealing previously hidden coordination failures and showing that \bench{} closely reflects real execution while significantly reducing evaluation time and token cost.

\end{itemize}

\begin{figure*}[!t]
\centering
\includegraphics[width=\textwidth]{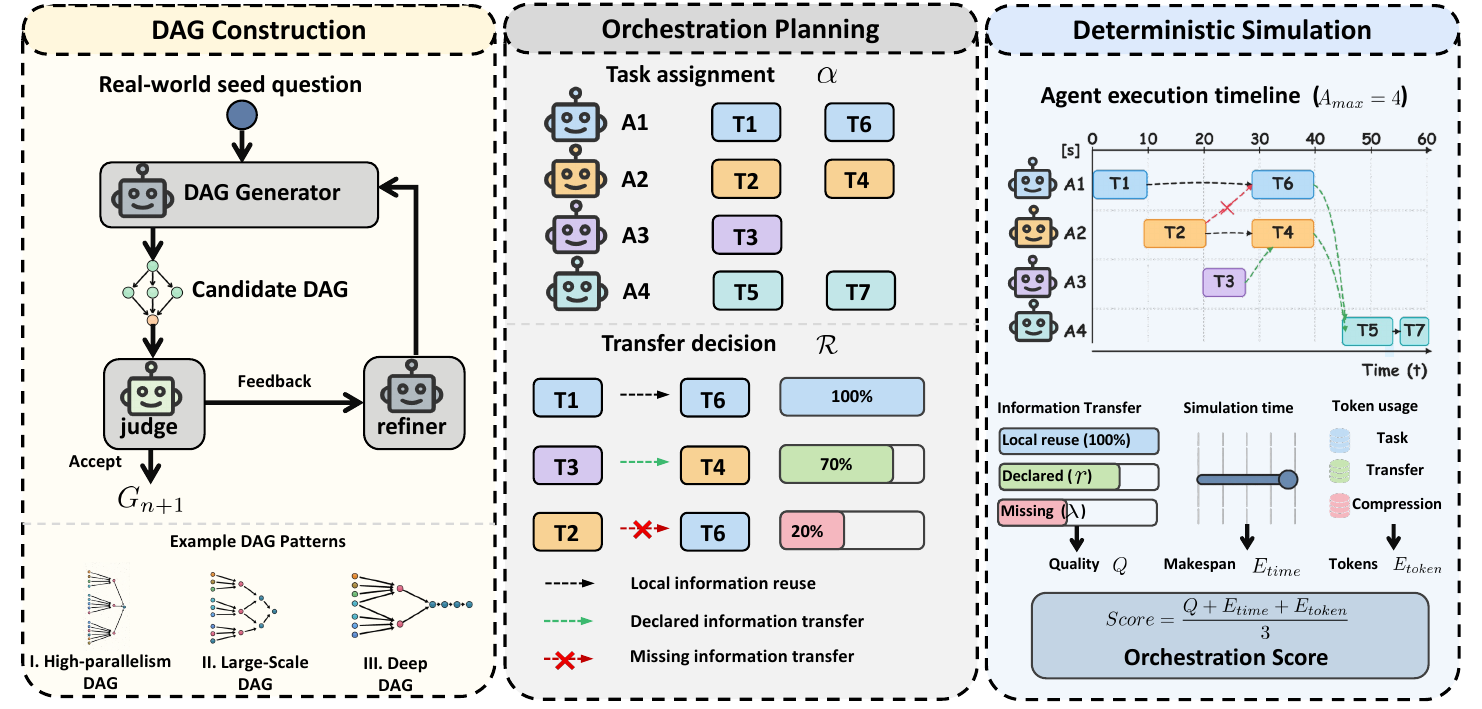}
\caption{Benchmark design overview. \bench{} first constructs a DAG from a raw seed question, then asks the planner model to produce an orchestration plan. The deterministic simulator executes the plan and obtains the score.}
\label{fig:benchmark-design-overview}
\end{figure*}

\section{Related Work}
\paragraph{End-to-End Evaluation of Agent Systems.} Most existing agent benchmarks test whether systems can complete tasks in external environments.
AgentBench covers web browsing, databases, operating systems, games, and embodied tasks \citep{liu2023agentbench}; GAIA evaluates multi-step assistant tasks requiring reasoning, information retrieval, and tool use \citep{mialon2023gaia}; and WebArena, SWE-bench, and OSWorld evaluate agents in realistic web, software-engineering, and computer-use environments \citep{zhou2023webarena,jimenez2023swebench,xie2024osworld}. ToolLLM studies large-scale API use \citep{qin2023toolllm}, while $\tau$-bench evaluates agents that interact with simulated users and domain-specific tools \citep{yao2024taubench}. MultiAgentBench extends end-to-end evaluation to multi-agent systems by measuring task completion and collaboration quality under different coordination protocols \citep{zhu2025multiagentbench}. MAST analyzes execution traces from multiple MAS frameworks and identifies system design, inter-agent misalignment, and task verification as major failure sources \citep{cemri2026multi}. These benchmarks are useful for evaluating deployed systems, but their results mix the effects of planning, worker performance, tools, and environments. It is therefore hard to tell whether success or failure comes from the orchestration plan.

\paragraph{Evaluation of Planning and Multi-Agent Orchestration.} Existing studies fall into three groups.  
The first evaluates planning and workflow structure. 
PlanBench tests formal action planning, FlowBench evaluates workflow-guided tool planning, and WorFBench compares generated workflows with reference sequences or graphs \citep{valmeekam2023planbench,xiao2024flowbench,qiao2025benchmarking}. Unlike \bench{}, they test structural correctness but do not simulate execution after tasks are assigned to agents.
The second group evaluates specific aspects of MAS orchestration. 
MASBENCH studies when MAS execution outperforms a single agent, PerspectiveGap evaluates information and prompt assignment across roles, and OrchRM focuses on MAS training \citep{ke2026mas,sun2026perspectivegap,tsang2026reward}. 
In contrast, \bench{} evaluates a complete orchestration plan without executing workers.
The third group studies distributed information use. 
HiddenBench and Silo-Bench test whether agents can communicate and combine information from separate contexts \citep{li2025assessing,zhang2026silo}. They evaluate coordination during interaction, whereas \bench{} evaluates task assignment, dependency scheduling, information transfer, and context compression beforehand.


\section{Problem Formulation}
In this work, we study orchestration planning under fixed task dependencies. 
Given a task DAG and resource constraints, the planner assigns subtasks to agents and specifies how required dependency information is transferred and retained across agents. 
Task decomposition and dependency structure are treated as fixed inputs rather than orchestration decisions.

Let $\mathcal{D}$ denote the pool of raw tasks.
Each generated task instance consists of a directed acyclic graph $G=(V,E)$ and a per-agent context limit $L$. Each node $v\in V$ represents a subtask, and each edge $(u,v)\in E$ indicates that $v$ requires information produced by $u$. The number of subtasks is $n=|V|$. Each subtask $v$ is associated with a description $d_v$, input, execution, and output token budgets $(x^{\mathrm{in}}_v,x^{\mathrm{exec}}_v,x^{\mathrm{out}}_v)$, a time budget $\tau_v$, and a compression-sensitivity class $c_v\in\{\mathrm{robust},\mathrm{balanced},\mathrm{fragile}\}$.
$L$ specifies the maximum number of tokens that each agent can retain in its working context. The task decomposition and dependency graph $G$ remain fixed throughout evaluation.

Given $G$, $L$, and a maximum agent budget $A_{\max}$, the
evaluated planner produces an orchestration plan
$\pi=(\alpha,\mathcal{R})$. 
The assignment function $\alpha:V\rightarrow\{1,\ldots,A_{\max}\}$
specifies which agent executes each subtask, while $\mathcal{R}$
specifies cross-agent information transfers. For a dependency edge
$(u,v)\in E$, if $\alpha(u)=\alpha(v)$, the output of $u$ is reused locally. If $\alpha(u)\neq\alpha(v)$ and the plan specifies $\mathcal{R}(u,v)=e$, where $e\in(0,1]$, the output of $u$ is transferred to the agent executing $v$ with retention ratio $e$.
A required cross-agent transfer omitted by the plan is recorded as a missing transfer, whereas a declared transfer that does not correspond to an edge in $E$ is treated as invalid. A deterministic simulator
evaluates $\pi$ under the task and context constraints and returns the
final result quality $Q$, makespan $M$, and total token cost $T$.

\section{Methodology}
\paragraph{Overview.}

As shown in Figure~\ref{fig:benchmark-design-overview}, \bench{} has
three stages: first, real-world seed tasks are transformed into dependency
DAGs. Given a DAG, context limit,
and agent budget, the planner produces the orchestration plan. 
The simulator
then evaluates the plan and produces an orchestration score.

\subsection{DAG Construction}

DAG Construction turns a suitable seed task into a DAG that requires multi-agent coordination.
We measure its natural parallelism as \(k_{99}/n\), where
\[
k_{99}(G)=\min\{k:\mathrm{ms}_k(G)\le C(G)+0.01(W(G)-C(G))\}.
\]
Here, \(W(G)\) is the serial execution time of all subtasks, \(C(G)\) is the critical-path time, and \(\mathrm{ms}_k(G)\) is the makespan with \(k\) workers.
Thus, \(k_{99}\) is the fewest workers needed to achieve \(99\%\) of the maximum possible reduction in makespan.
A larger \(k_{99}/n\) indicates greater natural parallelism.
We sample a seed task \(s\) from the problem pool \(\mathcal{D}\) and remove unsuitable tasks using a seed judge.
A refiner then constructs a DAG with a target number of subtasks \(n\) and a target \(k_{99}/n\).
In each round, one judge checks the format, while another checks the task decomposition and parallelism.
The refiner revises the DAG until both judges approve it.
The detailed process is provided in Appendix~E-II.

\subsection{Orchestration Planning}

Given the DAG \(G\), context limit \(L\), and maximum agent budget \(A_{\max}\), the tested planner \(F_\theta\) produces a workflow script \(z\). 
A deterministic interpreter then expands \(z\) as the concrete plan \(\pi=(\alpha,\mathcal{R})\).
An example plan is provided in Appendix~B-I.
The assignment \(\alpha\) maps each subtask to an agent, and a subtask
can start only after all of its dependencies have finished.
When a dependency edge \((u,v)\) connects tasks assigned to different agents, the plan must include an information transfer \(\mathcal{R}(u,v)=e\).
Its retention ratio \(e\in(0,1]\) specifies how much of the output of upstream task \(u\) is retained for downstream task \(v\). 
A larger \(e\) preserves more information but increases communication and context usage. 
When the context limit \(L\) of every agent is tight, the planner can reduce \(e\) to save space. 
Because subtasks differ in their sensitivity to information loss, robust outputs can tolerate stronger compression, whereas fragile outputs require higher retention. The planner must therefore choose task assignments and transfer ratios jointly to balance information quality, communication cost, and context usage.

\subsection{Deterministic Simulation}

The simulator draws on the six-stage lifecycle used by real execution frameworks: dependency resolution, agent scheduling, context acquisition, context management, subtask execution, and state updates. 
Given \(G\), \(\pi=(\alpha,\mathcal{R})\), and \(L\), the simulator
initializes each agent's clock and memory and processes subtasks in
topological order. For each subtask \(v\), it determines the start time,
collects and manages the required context, computes the output quality
and finish time, and updates the agent state. The updated state is then
used by downstream subtasks. After all subtasks finish, the simulator
computes the final quality, makespan, and token usage.
This alignment helps the simulator capture the execution consequences of each plan.
Please refer to Appendix~B-II for the detailed simulation process.

\paragraph{Dependency resolution}
In real execution, a subtask can begin only after all its dependencies are complete. 
Specifically, we preserve this by ensuring that every $u \in \text{Parents}(v)$ has finished before subtask v starts.
This does not mean that all tasks are serialized; in fact, independent subtasks can be assigned to different agents to achieve parallelism.

\paragraph{Agent scheduling}
In real execution, a subtask must also wait until its assigned agent is available. The simulator maintains a logical clock \(c_a\) for each agent. For \(v\) assigned to \(a=\alpha(v)\), its start time is
\begin{equation}
s_v=
\max\left(
c_a,\,
\max_{u\in\operatorname{Parents}(v)} f_u
\right),
\end{equation}
where \(f_u\) is the finish time of parent \(u\). 
Tasks assigned to the same agent run sequentially, while different agents can work in parallel. If several tasks are ready on one agent, the simulator follows a fixed topological order of the DAG.

\paragraph{Context acquisition}
A subtask can use only the information available to its assigned agent in real execution. 
The simulator reconstructs the effective input of \(v\) by tracing how each parent result reaches \(a=\alpha(v)\). 
Results produced by the same agent are reused locally. 
For cross-agent dependencies, the simulator checks the transfer map \(\mathcal{R}\), applies the declared compression ratio, and records the communication tokens and retained quality. 
If transfer is missing, the simulator records a missing-transfer event and sets the parent's effective quality to \(\lambda q\), where \(\lambda \in [0,1]\) is a penalty factor.

\paragraph{Context management}

Real agents operate under finite context limits: retaining more information may improve quality but increases token cost and context pressure. 
The simulator maintains a local memory \(M_a\) for each agent \(a\). Before executing \(v\), its input and available parent results are added to \(M_a\); if \(|M_a|>L\), the stored results are compressed until they fit. 
Let  \(n_v\) be the context usage of the agent and \(q_v\) the retained quality of the current subtask. 
Compressing it gives
\begin{equation}
n'_v=\left\lceil e\,n_v\right\rceil,
\qquad
q'_v=\max\!\left(q_{\min},q_ve_{s_v}\right),
\end{equation}
where \(e\in(0,1]\) is the context retention ratio, \(s_v\) denotes the compression sensitivity of \(v\), and \(e_{s_v}\) denotes the quality retention ratio.
More fragile information requires gentler compression, but consumes more context, creating a trade-off situation for the model.

\paragraph{Subtask execution}
After dependencies, agent availability, and input context are resolved, the simulator executes \(v\) deterministically from its task metadata. For a non-root task, let \(\widetilde q_{uv}\) denote the quality of parent \(u\)'s package available to \(v\).
Let $\bar q$ denote the geometric mean of the available parent qualities.
The result quality of  \(v\) is
\begin{align}
q_v
&=
\operatorname{clip}_{[q_{\min},1]}
\left(\bar q e_{s_v}\right),
\label{eq:task-quality}
\end{align}
where $e_{s_v}$ is the compression ratio for subtask $v$, if compression is applied.
The result is clipped to \([q_{\min},1]\).


\paragraph{State updates}
After execution, the simulator finishes the subtask $v$ with quality score \(q_v\) and stores it in the local memory of agent \(\alpha(v)\). The finish time includes the task time and any post-execution context-compression delay:
\begin{equation}
f_v=s_v+\tau_v+\Delta_v^{\mathrm{post}},
\qquad
c_{\alpha(v)}\leftarrow f_v,
\end{equation}
where \(\Delta_v^{\mathrm{post}}\) is the delay caused by compression. The simulator then releases packages that no remaining subtask needs.


\section{Experiment Settings}

\paragraph{Data.}
We construct the problem pool \(\mathcal{D}\) from four source datasets: Finance Agent~\cite{bigeard2025finance}, DS-1000~\cite{lai2023ds}, Qasper~\cite{dasigi2021dataset}, and BIRD-SQL~\cite{li2023can}.  
Using problems sampled from \(\mathcal{D}\), we generate 240 task DAGs for the main experiments. 
To study scalability, we additionally construct larger DAGs with \(n\in\{200,500,1000\}\) nodes. 
We use MultiAgentBench~\cite{zhu2025multiagentbench}, an established benchmark for multi-agent systems, to evaluate the agreement between simulation metrics and real-world execution results.
Appendix E-I provides an overview of our DAG dataset and demonstrates that it is comprehensive and balanced.
Appendix~E-III analyzes DAG construction failure cases, and in Appendix~E-IV we conduct experiments to show that our DAGs are of high quality.

\paragraph{Other Settings.}
For benchmark validation, we evaluate six models: GLM-5.1,
DeepSeek-V4-Pro, DeepSeek-V4-Flash, Qwen3.6-35B-A3B, Kimi-K2.6, and
Doubao-Seed-2.0-Mini~\cite{
zeng2026glm,
deepseek2026deepseek,
2025Qwen3,
moonshotai2026kimik26,
seed2026seed20modelcard}.
Our main evaluation covers these six models and three
frontier models: Gemini-3.1-Pro-Preview, Claude-Opus-4.8, and
GPT-5.5~\cite{
googledeepmind2026gemini31promodelcard,
anthropic2026claudeopus48systemcard,
openai2026gpt55systemcard}.
For real-world execution experiments, we run Claude Code under the
dynamic-workflow setting~\cite{anthropic2025claudecode}.
For the hyperparameters, we set $A_{max}=100$ and $\lambda=0.5$ (Appendix~F-III).
Other detailed settings are provided in Appendix~G.


\begin{table}[t]
\centering
\small
\begin{tabular}{llcccc}
\toprule
Group & Metric & Pearson \(r\) & \(p\) & Spearman \(\rho\) & \(p\) \\
\midrule
Scale
& DA & 0.973 & 0.001 & 1.000 & 0.003 \\
& CA & 0.749 & 0.090 & 0.829 & 0.058 \\
& SA & 0.829 & 0.056 & 0.714 & 0.136 \\
\midrule
Structure
& SDT & 0.928 & 0.011 & 0.943 & 0.017 \\
& PU & 0.768 & 0.035 & 1.000 & 0.003 \\
& WD & 0.887 & 0.028 & 0.771 & 0.103 \\
& ILR & 0.664 & 0.157 & 0.600 & 0.242 \\
\bottomrule
\end{tabular}
\caption{Simulation-to-real correlation on MultiAgentBench (\bench{} vs.\ Claude Code).}
\label{tab:real-run-orchestration-validation}
\end{table}

\begin{figure}[t]
\centering
\small
\begingroup
\newcommand{\corrCell}[4]{%
  \pgfmathsetmacro{\shade}{#4*85}
  \fill[teal!\shade!white] (#1,#2) rectangle ++(1,1);
  \draw[white,line width=0.4pt] (#1,#2) rectangle ++(1,1);
  \node at (#1+0.5,#2+0.5) {\scriptsize #3};
}

\begin{tikzpicture}[x=0.66cm,y=0.66cm]
\node[rotate=45,anchor=west] at (0.35,5.15) {\scriptsize \bench{}};
\node[rotate=45,anchor=west] at (1.35,5.15) {\scriptsize Claude Code};
\node[rotate=45,anchor=west] at (2.35,5.15) {\scriptsize SWE-mini};
\node[rotate=45,anchor=west] at (3.35,5.15) {\scriptsize OpenHands};
\node[rotate=45,anchor=west] at (4.35,5.15) {\scriptsize Crush};

\node[anchor=east] at (-0.1,4.5) {\scriptsize \bench{}};
\node[anchor=east] at (-0.1,3.5) {\scriptsize Claude Code};
\node[anchor=east] at (-0.1,2.5) {\scriptsize SWE-mini};
\node[anchor=east] at (-0.1,1.5) {\scriptsize OpenHands};
\node[anchor=east] at (-0.1,0.5) {\scriptsize Crush};

\corrCell{0}{4}{1.00}{1.00}
\corrCell{1}{3}{1.00}{1.00}
\corrCell{2}{2}{1.00}{1.00}
\corrCell{3}{1}{1.00}{1.00}
\corrCell{4}{0}{1.00}{1.00}

\corrCell{1}{4}{0.82}{0.82}
\corrCell{2}{4}{0.73}{0.73}
\corrCell{3}{4}{0.63}{0.63}
\corrCell{4}{4}{0.39}{0.39}
\corrCell{2}{3}{0.37}{0.37}
\corrCell{3}{3}{0.27}{0.27}
\corrCell{4}{3}{0.08}{0.08}
\corrCell{3}{2}{0.84}{0.84}
\corrCell{4}{2}{0.48}{0.48}
\corrCell{4}{1}{0.73}{0.73}

\corrCell{0}{3}{0.77}{0.77}
\corrCell{0}{2}{0.77}{0.77}
\corrCell{1}{2}{0.31}{0.31}
\corrCell{0}{1}{0.43}{0.43}
\corrCell{1}{1}{0.09}{0.09}
\corrCell{2}{1}{0.54}{0.54}
\corrCell{0}{0}{0.60}{0.60}
\corrCell{1}{0}{0.14}{0.14}
\corrCell{2}{0}{0.49}{0.49}
\corrCell{3}{0}{0.71}{0.71}

\end{tikzpicture}
\endgroup

\caption{Correlations between \bench{} and real quality across frameworks. Upper/lower triangles report Pearson/Spearman coefficients.}
\label{fig:framework-similarity}
\end{figure}

\paragraph{Metrics.}
We evaluate each execution plan in terms of output quality, scheduling
efficiency, and token efficiency. Let
\(\mathcal{T}=\{w\in V\mid\operatorname{outdeg}_{G}(w)=0\}\)
be the set of terminal tasks, and let \(q_w\in[0,1]\) be the
task-specific quality score of \(w\). Overall quality is the
macro-average over terminal tasks:
\begin{equation}
Q
=
\frac{1}{|\mathcal{T}|}
\sum_{w\in\mathcal{T}}q_w.
\label{eq:quality}
\end{equation}

Let \(C\) and \(M\) denote the weighted critical-path length and
observed makespan. Let \(T\) be the total token
consumption and \(T_{\mathrm{single}}\) that of the single-agent serial
baseline. We define
\begin{equation}
E_{\mathrm{time}}
=
\min\left\{1,\frac{C}{M}\right\},
\qquad
E_{\mathrm{token}}
=
\min\left\{1,\frac{T_{\mathrm{single}}}{T}\right\}.
\label{eq:efficiency}
\end{equation}
Here, \(E_{\mathrm{time}}\) measures proximity to the critical-path
lower bound, while \(E_{\mathrm{token}}\) measures token efficiency
relative to the serial baseline.

The final score is the arithmetic mean of the three metrics:
\begin{equation}
\mathrm{Score}
=
\frac{Q+E_{\mathrm{time}}+E_{\mathrm{token}}}{3}.
\label{eq:final_score}
\end{equation}

We also report makespan, token use, agent count, and missing transfers for diagnostic
analysis.



\begin{table}[t]
\centering
\small
\begin{tabular}{lcccc}
\toprule
Metric & Pearson \(r\) & \(p\) & Spearman \(\rho\) & \(p\) \\
\midrule
Final score & 0.816 & 0.047 & 0.771 & 0.103 \\
Time & -0.264 & 0.633 & -0.314 & 0.564 \\
Token usage & -0.607 & 0.206 & -0.371 & 0.497 \\
\bottomrule
\end{tabular}
\caption{Correlations between our metrics and real outcomes.}
\label{tab:real-run-global-validation}
\end{table}

\begin{table}[t]
\centering
\footnotesize
\setlength{\tabcolsep}{3.5pt}
\renewcommand{\arraystretch}{1.02}
\begin{tabular}{@{}lcccc@{}}
\toprule
& \multicolumn{2}{c}{Pearson}
& \multicolumn{2}{c}{Spearman} \\
\cmidrule(lr){2-3}
\cmidrule(lr){4-5}
Setting & \(r\) & \(p\) & \(\rho\) & \(p\) \\
\midrule
Original        & 0.816 & 0.047 & 0.771 & 0.103 \\
w/o DeepSeek-V4-Flash & 0.677 & 0.183 & 0.600 & 0.350 \\
w/o DeepSeek-V4-Pro   & 0.682 & 0.192 & 0.900 & 0.083 \\
w/o Doubao-Mini & 0.421 & 0.500 & 0.600 & 0.350 \\
w/o GLM-5.1     & 0.949 & 0.033 & 0.900 & 0.083 \\
w/o Kimi-K2.6   & 0.790 & 0.108 & 0.900 & 0.083 \\
w/o Qwen3.6-A3B & 0.719 & 0.167 & 0.600 & 0.350 \\
\bottomrule
\end{tabular}
\caption{Leave-one-model-out correlations between simulated scores and real task quality under Claude Code.}
\label{tab:leave-one-model-out-correlation}
\end{table}


\begin{table*}[t]
\centering
\begingroup
\scriptsize
\setlength{\tabcolsep}{2.8pt}
\renewcommand{\arraystretch}{0.94}

\begin{minipage}[t]{0.49\textwidth}
\centering
\textbf{\(n=10\)}
\vspace{2pt}

\begin{tabular*}{\linewidth}{@{\extracolsep{\fill}}lrrrrrr@{}}
\toprule
Model & \#Agents & Miss.$\downarrow$ & Token$\uparrow$ & Speed$\uparrow$ & Qual.$\uparrow$ & Score$\uparrow$ \\
\midrule
GPT-5.5              & 6.42 & 0.12 & 0.735 & \textbf{0.744} & 0.951 & \textbf{0.810} \\
Gemini-3.1-Pro       & 6.40 & \textbf{0.00} & 0.730 & 0.714 & \textbf{0.960} & 0.801 \\
GLM-5.1              & 6.40 & 0.43 & 0.740 & 0.735 & 0.917 & 0.797 \\
DeepSeek-V4-Pro      & 6.18 & 0.05 & 0.739 & 0.676 & 0.928 & 0.781 \\
Claude-Opus-4.8      & 6.48 & \textbf{0.00} & 0.721 & 0.656 & \textbf{0.960} & 0.779 \\
DeepSeek-V4-Flash    & 6.36 & 0.24 & 0.722 & 0.660 & 0.925 & 0.769 \\
Qwen3.6-35B-A3B      & 6.00 & 0.25 & 0.739 & 0.650 & 0.916 & 0.768 \\
Kimi-K2.6            & 5.85 & 0.43 & \textbf{0.743} & 0.647 & 0.914 & 0.768 \\
Doubao-Seed-2.0-Mini & 6.28 & 0.25 & 0.731 & 0.635 & 0.919 & 0.761 \\
\bottomrule
\end{tabular*}
\end{minipage}
\hfill
\begin{minipage}[t]{0.49\textwidth}
\centering
\textbf{\(n=20\) }
\vspace{2pt}

\begin{tabular*}{\linewidth}{@{\extracolsep{\fill}}lrrrrrr@{}}
\toprule
Model & \#Agents & Miss.$\downarrow$ & Token$\uparrow$ & Speed$\uparrow$ & Qual.$\uparrow$ & Score$\uparrow$ \\
\midrule
GLM-5.1              & 12.15 & 0.58 & 0.693 & \textbf{0.652} & 0.893 & \textbf{0.746} \\
Gemini-3.1-Pro       & 12.33 & \textbf{0.07} & 0.682 & 0.638 & \textbf{0.910} & 0.743 \\
GPT-5.5              & 11.80 & 0.75 & 0.691 & 0.643 & 0.876 & 0.737 \\
Claude-Opus-4.8      & 12.02 & 0.10 & 0.685 & 0.616 & 0.892 & 0.731 \\
DeepSeek-V4-Pro      & 11.74 & 0.79 & 0.674 & 0.612 & 0.826 & 0.704 \\
Doubao-Seed-2.0-Mini & 11.55 & 3.47 & 0.704 & 0.628 & 0.777 & 0.703 \\
Qwen3.6-35B-A3B      & 11.00 & 3.60 & \textbf{0.715} & 0.633 & 0.733 & 0.694 \\
Kimi-K2.6            & 10.72 & 2.97 & 0.689 & 0.593 & 0.756 & 0.679 \\
DeepSeek-V4-Flash    & 12.12 & 1.16 & 0.562 & 0.500 & 0.695 & 0.586 \\
\bottomrule
\end{tabular*}
\end{minipage}

\vspace{7pt}

\begin{minipage}[t]{0.49\textwidth}
\centering
\textbf{\(n=50\)}
\vspace{2pt}

\begin{tabular*}{\linewidth}{@{\extracolsep{\fill}}lrrrrrr@{}}
\toprule
Model & \#Agents & Miss.$\downarrow$ & Token$\uparrow$ & Speed$\uparrow$ & Qual.$\uparrow$ & Score$\uparrow$ \\
\midrule
Claude-Opus-4.8      & 29.65 & 0.43 & 0.635 & 0.484 & \textbf{0.813} & \textbf{0.644} \\
Gemini-3.1-Pro       & 31.02 & \textbf{0.00} & 0.627 & 0.494 & 0.804 & 0.642 \\
GPT-5.5              & 30.87 & 0.67 & 0.619 & 0.491 & 0.807 & 0.639 \\
GLM-5.1              & 29.48 & 4.42 & 0.654 & \textbf{0.503} & 0.755 & 0.637 \\
Qwen3.6-35B-A3B      & 29.68 & 6.45 & 0.652 & 0.485 & 0.647 & 0.595 \\
Kimi-K2.6            & 30.45 & 5.83 & 0.621 & 0.471 & 0.660 & 0.584 \\
DeepSeek-V4-Flash    & 29.64 & 3.55 & 0.603 & 0.447 & 0.647 & 0.566 \\
Doubao-Seed-2.0-Mini & 30.70 & 9.18 & \textbf{0.658} & 0.474 & 0.561 & 0.564 \\
DeepSeek-V4-Pro      & 31.17 & 1.43 & 0.564 & 0.420 & 0.662 & 0.548 \\
\bottomrule
\end{tabular*}
\end{minipage}
\hfill
\begin{minipage}[t]{0.49\textwidth}
\centering
\textbf{\(n=100\)}
\vspace{2pt}

\begin{tabular*}{\linewidth}{@{\extracolsep{\fill}}lrrrrrr@{}}
\toprule
Model & \#Agents & Miss.$\downarrow$ & Token$\uparrow$ & Speed$\uparrow$ & Qual.$\uparrow$ & Score$\uparrow$ \\
\midrule
Gemini-3.1-Pro       & 63.23 & \textbf{0.07}  & 0.614 & \textbf{0.414} & \textbf{0.690} & \textbf{0.573} \\
GLM-5.1              & 63.50 & 5.53  & 0.629 & 0.410 & 0.651 & 0.563 \\
GPT-5.5              & 62.40 & 6.32  & 0.611 & \textbf{0.414} & 0.650 & 0.558 \\
DeepSeek-V4-Pro      & 60.72 & 7.58  & 0.631 & 0.401 & 0.609 & 0.547 \\
Claude-Opus-4.8      & 62.61 & 0.75  & 0.605 & 0.388 & 0.639 & 0.544 \\
Qwen3.6-35B-A3B      & 58.23 & 14.37 & \textbf{0.645} & 0.406 & 0.492 & 0.514 \\
DeepSeek-V4-Flash    & 59.65 & 4.53  & 0.599 & 0.379 & 0.564 & 0.514 \\
Doubao-Seed-2.0-Mini & 63.80 & 22.70 & 0.644 & 0.394 & 0.443 & 0.493 \\
Kimi-K2.6            & 59.07 & 13.00 & 0.585 & 0.364 & 0.508 & 0.485 \\
\bottomrule
\end{tabular*}
\end{minipage}

\endgroup
\caption{Main results across DAG sizes. Each panel reports results for DAGs with \(n\) tasks.}
\label{tab:main-results}
\end{table*}

\section{Results and Discussion}
We first establish the fidelity of our simulator by demonstrating strong agreement between simulated metrics and real executions.
Using the validated simulator, we then study multi-agent orchestration at scale. 
Our results show that preserving task-critical information is more important than simply increasing the number of agents: as coordination failures accumulate, parallel execution yields diminishing returns. 
Ablation studies further support our transfer and compression designs, with additional results provided in Appendix~A.

\subsection{Benchmark Validation}

\noindent\textbf{Structural alignment.}
Table~\ref{tab:real-run-orchestration-validation} reports model-level correlations between simulated orchestration metrics and the corresponding metrics from real Claude Code workflows on MultiAgentBench.
For scale, DA measures agreement in the number of declared subagents, SA in the number of launched subagents, and CA in the number of subagents that return results. 
For structure, SDT, PU, WD, and ILR measure agreement in delegation tendency, subagent
parallelism, dependency depth, and missing information transfers, respectively. 
The correlations across both categories indicate that \bench{} captures key orchestration behaviors.
The strongest agreement appears in both scale and structure, demonstrating consistent alignment with real-world orchestration patterns.
Detailed metric definitions are provided in Appendix~D-I.

\noindent\textbf{Outcome alignment.}
Table~\ref{tab:real-run-global-validation} shows that simulated scores strongly correlate with real task quality (Pearson \(r=0.816\); Spearman \(\rho=0.771\)). This correlation remains positive in all leave-one-model-out settings and reaches \(r=0.949\) without GLM-5.1 (Table~\ref{tab:leave-one-model-out-correlation}). We assign each subtask time and token costs during DAG generation.
These controlled costs test how planners balance quality and efficiency, making the time and token scores meaningful for planner comparison. 
However, real time and token consumption are framework-dependent and cannot be reliably predicted (Appendix~D-II), which explains their lower simulation-to-real correlations.


\noindent\textbf{Cross-framework robustness.}
Figure~\ref{fig:framework-similarity} evaluates the agreement between \bench{} simulations and real executions using four agent frameworks:
Claude Code~\cite{anthropic2025claudecode},
SWE-mini~\cite{yang2024sweagent},
OpenHands~\cite{2024OpenHands}, and
Crush~\cite{charmbracelet2025crush}.
\bench{} maintains 
positive correlations across all four frameworks,
indicating that its predictions are not tied to a particular implementation. 
These results support using \bench{} for initial screening, followed by validation in the target framework.




\subsection{Evaluating Orchestration}
We evaluate the six validation models together with GPT-5.5,
Gemini-3.1-Pro-Preview, and Claude-Opus-4.8 on DAGs with
\(n\in\{10,20,50,100\}\) tasks. 
In Table~\ref{tab:main-results}, \#Agents and Miss.\ denote the average numbers of declared agents and missing information transfers, respectively.
Token, Speed, Qual., and Score denote token efficiency, scheduling efficiency, output quality, and the aggregate score defined in
Equation~\ref{eq:final_score}.
As the workflows grow, information routing becomes increasingly
challenging: the range of average missing transfers widens from
\([0.00,0.43]\) at \(n=10\) to \([0.07,22.70]\) at \(n=100\).
No model achieves the highest aggregate score at every scale:
GPT-5.5 leads at \(n=10\), GLM-5.1 at \(n=20\), Claude-Opus-4.8 at \(n=50\), and Gemini-3.1-Pro-Preview at \(n=100\).
Additional analysis is provided in Appendix~F-I.

\paragraph{Transfer Coverage Matters More Than Agent Count.}
Table~\ref{tab:coverage-quality-correlation} reports correlations between agent count, transfer coverage, and performance. Agents--\(Q\) and Agents--Score measure the effect of agent count on quality score and the final score, while Coverage--\(Q\) measures the effect of successful transfers. 
Agent count is nearly uncorrelated with quality (\(-0.021\)) at $n=100$, whereas Coverage is consistently more informative across scales (\(0.614\)--\(0.952\)), showing that more agents do not necessarily improve orchestration.


\begin{figure}[t]
    \centering
    \includegraphics[width=\linewidth]{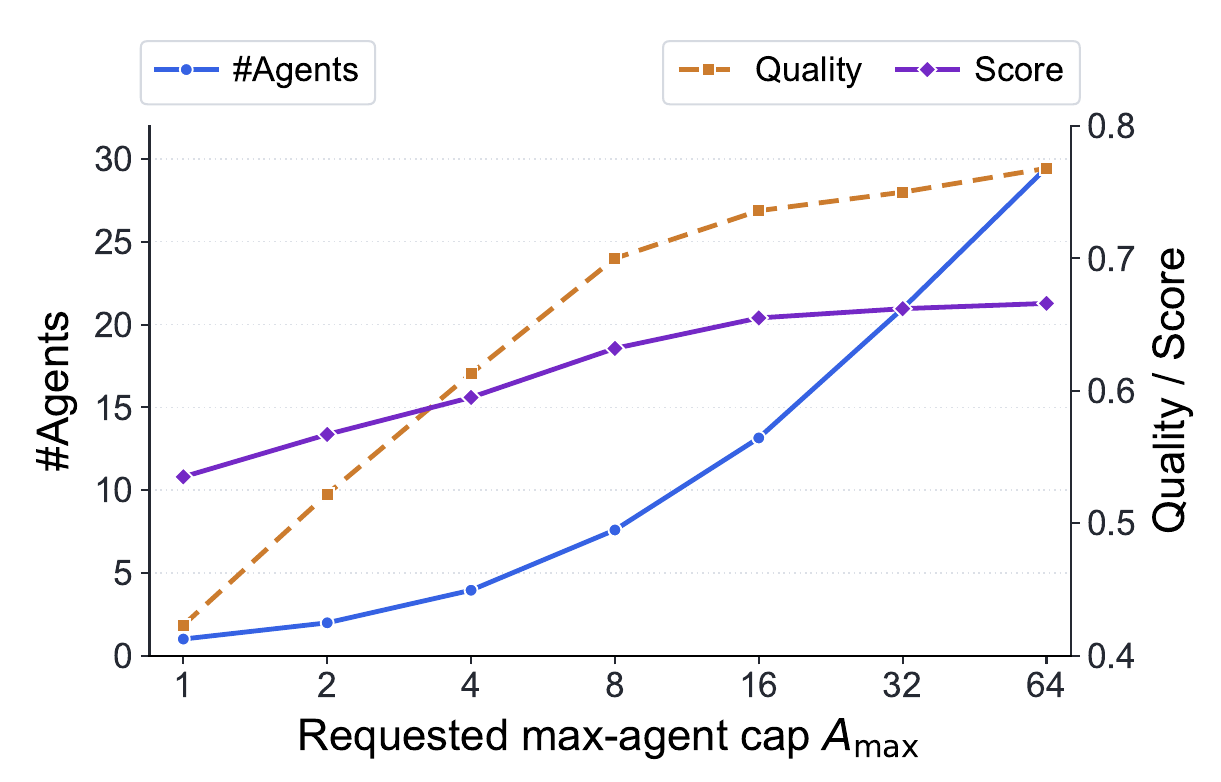}
    \caption{
    Max-agent-cap sweep averaged across the six base planners. 
    }
    \label{fig:max-agent-sweep}
\end{figure}

\begin{figure*}[t]
    \centering
    \includegraphics[
        width=\textwidth
    ]{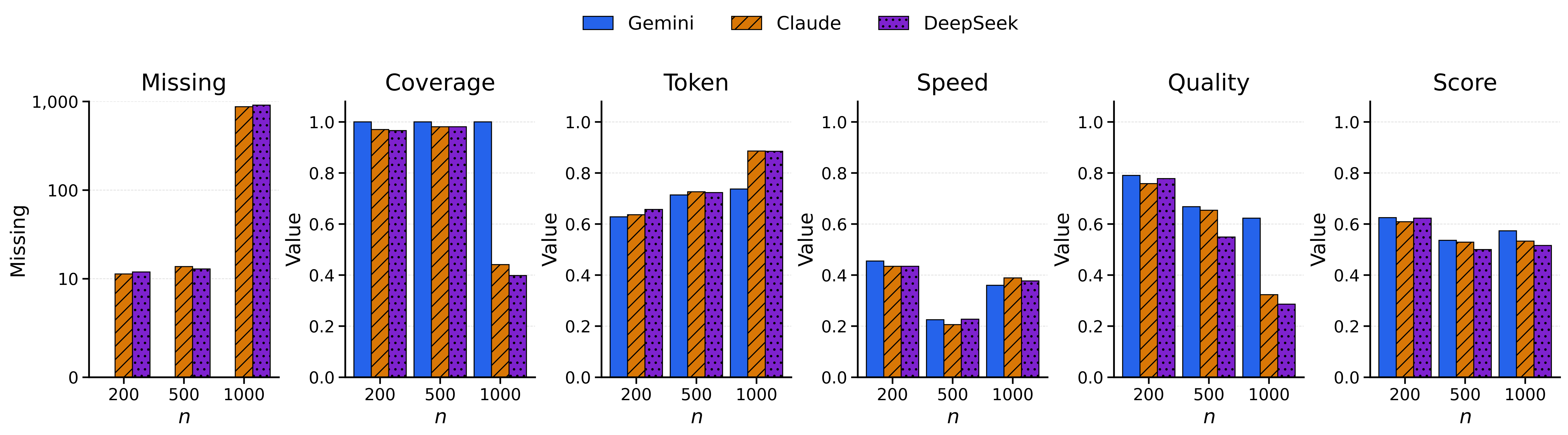}
    \caption{ Extreme-scale results under \(A_{\max}=100\).}
    \label{fig:coverage-quality-score}
\end{figure*}

\begin{table}[t]
\centering
\footnotesize
\setlength{\tabcolsep}{3pt}
\renewcommand{\arraystretch}{1.05}
\begin{tabular}{@{}lrrr@{}}
\toprule
n & Agents--\(Q\) & Coverage--\(Q\) & Agents--Score \\
\midrule
10  &  0.468 & 0.952 & -0.481 \\
20  &  0.720 & 0.807 &  0.218 \\
50  &  0.470 & 0.829 & -0.189 \\
100 & -0.021 & 0.614 & -0.676 \\
\bottomrule
\end{tabular}
\caption{Pearson correlations between orchestration diagnostics and scores. \(Q\) denotes quality.}
\label{tab:coverage-quality-correlation}
\end{table}

\begin{table}[t]
\centering
\footnotesize
\setlength{\tabcolsep}{4pt}
\renewcommand{\arraystretch}{1.02}
\begin{tabular}{@{}lrrrrr@{}}
\toprule
& \multicolumn{3}{c}{Quality} & \multicolumn{2}{c}{Final score} \\
\cmidrule(lr){2-4}
\cmidrule(lr){5-6}
$L$ & Single & Multi & \(\Delta\) & Single & Multi \\
\midrule
\(16\mathrm{k}\)  & 0.423 & 0.725 & \(+0.302\) & 0.535 & 0.650 \\
\(32\mathrm{k}\)  & 0.649 & 0.821 & \(+0.172\) & 0.613 & 0.684 \\
\(64\mathrm{k}\)  & 0.792 & 0.852 & \(+0.060\) & 0.662 & 0.692 \\
\(128\mathrm{k}\) & 0.852 & 0.859 & \(+0.007\) & 0.682 & 0.693 \\
\bottomrule
\end{tabular}
\caption{Single/multi-agent performance under different $L$.}
\label{tab:context-sweep-small}
\end{table}

\paragraph{Orchestration Reliability Diverges at Scale.}
\label{sec: Scaling}
To test how this information-flow bottleneck scales, we evaluate Gemini-3.1-Pro-Preview, Claude-Opus-4.8, and DeepSeek-V4-Flash on DAGs with \(n=200\), \(n=500\), and \(n=1{,}000\), under \(A_{\max}=100\). 
As shown in Figure~\ref{fig:coverage-quality-score}, from \(500\) to \(1{,}000\) tasks, transfer coverage falls from \(0.981\) to \(0.441\) for Claude and from \(0.981\) to \(0.398\) for DeepSeek, producing \(872.3\) and \(907.4\) missing transfers. Gemini retains complete coverage and substantially higher quality. 
The degradation reflects coordination failures that accumulate as the numbers of dependencies and handoffs increase.
\paragraph{More Agents Yield Diminishing Returns.}
We next ask whether increasing the agent budget improves orchestration.
Figure~\ref{fig:max-agent-sweep} summarizes the results
across six base planners and 50 DAGs.
A larger \(A_{\max}\) initially reduces compression pressure and improves quality and score. 
However, these gains quickly saturate: increasing \(A_{\max}\) from \(16\) to \(64\) more than doubles agent count while leaving scores nearly unchanged.


\subsubsection{When Do More Agents Help?}
We also compare MAS and single-agent execution on the same 50 DAGs to show when a single agent outperforms the MAS. 
Table~\ref{tab:context-sweep-small} shows that at \(L=16\mathrm{k}\), multi-agent execution yields a 0.302 improvement in quality by avoiding repeated single-agent compression. 
The advantage falls to \(0.007\) at \(128\mathrm{k}\), where single-agent quality is already higher on the 10-, 20-, and 50-task DAGs. 
Additional agents are therefore most useful when the working state exceeds one context window; once it fits, coordination can become pure overhead.
Detailed per-model performance results can be found in Appendix~F-II.

\begin{table}[t]
\centering
\small
\setlength{\tabcolsep}{5pt}
\renewcommand{\arraystretch}{1.02}
\begin{tabular}{@{}lcc@{}}
\toprule
 Selection & Spearman & Pair accuracy \\
\midrule
\bench{} & \textbf{0.754} & \textbf{0.800} \\
Random          & 0.176 & 0.613 \\
\bottomrule
\end{tabular}
\caption{Model-ranking agreement under limited real-execution budgets for five tasks.}
\label{tab:low-budget-task-selection}
\end{table}

\begin{table}[t]
\centering
\small
\setlength{\tabcolsep}{4pt}
\renewcommand{\arraystretch}{1.05}
\begin{tabular}{lcc}
\toprule
Benchmark
& Tokens (real/ours)
& Time (real/ours) \\
\midrule
WideSearch
& \(17.35\mathrm{M}/44.61\mathrm{K}\)
& \(56.13/0.51\) min \\
MultiAgentBench
& \(383.09\mathrm{K}/5.16\mathrm{K}\)
& \(5.61/0.58\) min \\
\bottomrule
\end{tabular}
\caption{Average resource consumption per task.}
\label{tab:resource-consumption}
\end{table}

\begin{table}[t]
\centering
\small
\begin{tabular}{llrr}
\toprule
Evaluation & Metric & Baseline $A$ & Refined $B$ \\
\midrule
\multirow{4}{*}{Simulation}
& Quality & 0.1452 & \textbf{0.5111} \\
& Speed   & 0.9231 & \textbf{0.9163} \\
& Token   & 0.6858 & \textbf{0.6421} \\
& Missing & 6.80   & \textbf{5.85}   \\
\midrule
Real execution
& Score & 3.754 & \textbf{4.150} \\
\midrule
\end{tabular}
\caption{
Simulator-guided workflow refinement results.
}
\label{tab:simulator-guided-handoff}
\end{table}

\paragraph{Simulation Enables Cost-Efficient Evaluation and Selection.}
Beyond diagnosis, \bench{} helps allocate limited real-execution budgets.
We rank tasks by cross-model
disagreement in their simulated scores and execute only the five tasks
with the highest disagreement.
As shown in Table~\ref{tab:low-budget-task-selection}, our method achieves a Spearman correlation of \(0.754\) and pairwise accuracy of \(0.800\), surpassing random selection and substantially reducing the number of executions needed for model comparison.
We also demonstrate the use of task-difficulty estimation and model selection in Appendix~C.


\paragraph{Simulation Cuts Token and Runtime by at least \(74\times\) and \(9.7\times\).}
Our approach is practical because simulation is substantially cheaper than real execution. Table~\ref{tab:resource-consumption} shows reductions of \(389\times\) in token and \(110\times\) in time on WideSearch, and \(74\times\) and \(9.7\times\) on MultiAgentBench. This efficiency enables cost-effective plan screening, large-scale stress testing, and controlled parameter sweeps before targeted real executions.

\paragraph{Simulator-guided Refinement Improves Quality.}
We evaluate simulator-guided workflow refinement on 20 MultiAgentBench tasks, using DeepSeek-V4-Flash as the execution model and DeepSeek-V4-Pro as the evaluator.
For each baseline workflow ($A$), the refined workflow ($B$) adds one simulator-selected cross-role handoff while keeping all other components fixed; if no handoff improves the simulation, $B=A$.
As shown in Table~\ref{tab:simulator-guided-handoff}, the mean real-execution score increases from 3.754 to 4.150 out of \(5\).

\section{Conclusion}

We introduced OrchBench, a benchmark for evaluating multi-agent orchestration through lightweight simulation. 
Our results show that good orchestration depends less on the number of agents than on preserving information across dependent tasks. 
Additional agents help relieve context pressure, but also introduce diminishing returns and coordination failures. 
The simulated results also correlate with real execution, supporting OrchBench as an efficient tool for screening orchestration plans before framework-specific validation.
\bibliography{aaai2027}

\newpage
\appendix

\begin{table*}[t]
\centering
\caption{Mechanism ablations for Gemini and Doubao. Gaps are Gemini minus Doubao.}
\label{tab:mechanism-ablation}
\small
\begin{tabular}{lrrrrrr}
\toprule
Setting
& Gemini Q & Doubao Q & \(\Delta Q\)
& Gemini Score & Doubao Score & \(\Delta\)Score \\
\midrule
Full benchmark
& 0.690 & 0.443 & 0.247
& 0.573 & 0.493 & 0.079 \\
No missing penalty
& 0.698 & 0.686 & 0.011
& 0.575 & 0.575 & 0.001 \\
Auto-completion
& 0.697 & 0.663 & 0.035
& 0.575 & 0.555 & 0.020 \\
Auto-completion + lossless
& 0.933 & 0.933 & $-0.001$
& 0.653 & 0.645 & 0.008 \\
\bottomrule
\end{tabular}
\end{table*}

\section{A. Ablation Study}
\label{sec:ablation_study}

To validate the effectiveness of the mechanisms introduced in our benchmark, we compare Gemini~3.1~Pro and Doubao~Seed~2.0~Mini under controlled ablations. The two models represent a strong and a weak planner, respectively. If removing the proposed mechanisms makes their performance nearly indistinguishable, the benchmark loses its ability to discriminate orchestration capability.

We replay the same valid plans while keeping the DAGs, task assignments, agent budgets, and context limits fixed. Quality \(Q\) measures the information quality retained at the terminal tasks after cross-agent transfer and context compression. Score is the equal-weight average of quality, speed, and token efficiency. We report
\(\Delta Q=Q_{\text{Gemini}}-Q_{\text{Doubao}}\) and
\(\Delta\mathrm{Score}=\mathrm{Score}_{\text{Gemini}}-\mathrm{Score}_{\text{Doubao}}\)
as direct measures of model separation.

In Table~\ref{tab:mechanism-ablation}, the \emph{no missing penalty} setting removes the quality penalty for an omitted cross-agent transfer without introducing additional time or token costs. The \emph{auto-completion} setting instead inserts a normal handoff for every omitted cross-agent dependency and charges its communication tokens, handoff time, and context cost. The combined setting applies auto-completion and additionally makes context compression lossless: compression still reduces context size and incurs its normal time and token costs, but it no longer reduces package quality.

Under the full benchmark, Gemini outperforms Doubao by \(0.247\) in quality and \(0.079\) in score. Removing the missing-transfer penalty reduces these gaps to \(0.011\) and \(0.001\), while automatically completing omitted transfers reduces them to \(0.035\) and \(0.020\). This shows that transfer coverage is a major source of model separation.

The combined ablation further amplifies this effect. When omitted transfers are automatically completed and context compression becomes lossless, both models obtain a quality score of approximately \(0.933\). The quality gap becomes effectively zero, and the score gap falls to \(0.008\). Thus, once both transfer omissions and context-compression losses are neutralized, the strong and weak planners become nearly indistinguishable. This result confirms that the two mechanisms jointly prevent score saturation and preserve the benchmark's discriminative power.

\begin{algorithm}[!t]
\caption{Deterministic simulation of an orchestration plan}
\label{alg:simulation}
\small
\begin{algorithmic}[1]
\Statex \textbf{Input:} problem package \(P=(G,L)\), concrete plan
\(\pi=(\alpha,\mathcal{R})\), max-agent cap \(A_{\max}\),
missing-transfer factor \(\lambda\)
\Statex \textbf{Each subtask has:} input tokens, execution tokens,
result tokens, time cost, and compression class.
\Statex \textbf{Each result package has:} a token size and a quality value.

\If{\(\pi\) misses tasks, uses too many agents, has invalid task IDs, or has illegal transfers}
    \State \Return \(\mathrm{ZeroScore}(\mathrm{invalid\mbox{-}plan})\)
\EndIf

\State initialize each agent's clock and memory
\State initialize total tokens, finish times, and result qualities

\ForAll{\(v\in\mathrm{TopologicalOrder}(G)\)}
    \State \(a\gets\) the agent assigned to subtask \(v\)
    \State wait until all parent subtasks of \(v\) are finished and agent \(a\) is idle

    \ForAll{\(u\in\mathrm{Parents}(v)\)}
        \If{\(u\) and \(v\) are assigned to the same agent}
            \State reuse \(u\)'s result with no transfer cost or quality loss
        \ElsIf{\(\mathcal{R}\) declares a transfer from \(u\) to \(v\)}
            \State copy \(u\)'s result into agent \(a\)'s memory using the declared compression ratio
            \State add transfer tokens and record the retained quality
        \Else
            \State record a missing-transfer event
            \State multiply \(u\)'s contribution quality by \(\lambda\)
        \EndIf
    \EndFor

    \State add \(v\)'s input context to agent \(a\)'s memory and token count
    \If{agent \(a\)'s memory exceeds \(L\)}
        \State compress stored packages until the memory fits
        \State add compression tokens and record the quality loss
    \EndIf

    \State execute \(v\); add execution tokens, result tokens, and time cost
    \State compute \(v\)'s result quality from the available parent information
    \State store \(v\)'s result package in agent \(a\)'s memory

    \If{agent \(a\)'s memory exceeds \(L\)}
        \State compress stored packages until the memory fits
        \State add compression tokens and record the quality loss
    \EndIf

    \State record \(v\)'s finish time
    \State release parent packages that no future subtask needs
\EndFor

\State \Return final quality, makespan, total tokens, and diagnostics
\end{algorithmic}
\end{algorithm}

\begin{figure}[t]
\centering
\begin{lstlisting}[style=workflowjson]
{
  "agent_pools": [
    {"name": "workers", "count": 2},
    {"name": "final", "count": 1}
  ],
  "rules": [
    {
      "match": {"stage": "Analysis"},
      "pool": "workers",
      "strategy": "round_robin"
    },
    {
      "match": {"stage": "Synthesis"},
      "pool": "final",
      "strategy": "dependency_locality"
    }
  ],
  "default": {
    "pool": "workers",
    "strategy": "load_balance"
  },
  "transfer_rules": [
    {
      "parent_match": {"stage": "Analysis"},
      "child_match": {"stage": "Synthesis"},
      "cross_agent_only": true,
      "compression": 0.80
    }
  ]
}
\end{lstlisting}
\caption{An example workflow script.}
\label{fig:workflow-script-example}
\end{figure}

\section{B. Planning and Simulation Process}
\label{sec:Planning_simulation_process}

\subsection{I. Planning instance}
At the planning stage, each model outputs a JSON-formatted
\texttt{workflow\_script} that declaratively specifies how the input DAG
should be executed. The script defines agent pools, ordered task-matching
rules, a default rule for unmatched tasks, and dependency-based transfer
rules. 
The matching rules determine which pool and assignment strategy are
used for each task, whereas the transfer rules specify which intermediate
results should be sent across agents and how much information is retained
after compression. 
The script does not contain executable code or enumerate
every task assignment explicitly. 
Instead, the benchmark interpreter applies
the rules in topological order and deterministically expands the compact
script into concrete agent assignments and transfer steps. 
Figure~\ref{fig:workflow-script-example}
shows an example workflow script.

\subsection{II. Simulation Process}

The simulator receives the problem package \(P=(G,L)\) and the concrete orchestration plan \(\pi=(\alpha,\mathcal{R})\). It first checks whether the plan covers every subtask, respects the maximum agent budget, and contains only valid assignments and transfers. Invalid plans receive zero score. After this check, the simulator executes the plan without calling the planner model again.

The simulator processes subtasks in a deterministic topological order. For each subtask \(v\), its assigned agent waits until all parent subtasks have finished and the agent itself is idle. If a parent \(u\) is assigned to the same agent, its result is reused directly without communication. If \(u\) is assigned to another agent and the plan declares a transfer, the simulator copies the result using the specified compression ratio and records the communication cost and retained quality. If the required transfer is missing, the simulator records a missing-transfer event and reduces the contribution of \(u\) by the factor \(\lambda\).

Before executing \(v\), the simulator adds its input context to the assigned agent's memory. If the memory exceeds \(L\), stored packages are compressed until they fit, incurring additional token cost and information loss. The simulator then executes \(v\), charges its token and time costs, computes its result quality from the available parent information, and stores the resulting package in the agent's memory. The context limit is checked again after execution, and parent packages are released once they are no longer needed.

After all subtasks have completed, the simulator returns the final quality, makespan, total token cost, and diagnostic events. Because the execution order and all state updates follow fixed rules, the same problem and orchestration plan always produce the same result.

\begin{table*}[t]
\centering
\small
\caption{Utility of \bench{} for model selection on MultiAgentBench.}
\label{tab:utility-model-selection}
\resizebox{0.9\linewidth}{!}{
\begin{tabular}{llcccccc}
\toprule
Strategy & Scenario & Top-1 Cov. & Top-1 Reg. & Top-2 Cov. & Top-2 Reg. & Top-3 Cov. & Top-3 Reg. \\
\midrule
OrchBench & Overall & 61.5\% & 0.071 & 74.4\% & 0.033 & 81.8\% & 0.024 \\
Historical best & Overall & 38.6\% & 0.142 & 73.3\% & 0.053 & 82.2\% & 0.023 \\
OrchBench & Negotiation & 66.7\% & 0.067 & 86.7\% & 0.027 & 93.3\% & 0.009 \\
Historical best & Negotiation & 46.7\% & 0.120 & 93.3\% & 0.009 & 93.3\% & 0.004 \\
OrchBench & Coding & 50.0\% & 0.090 & 61.5\% & 0.038 & 57.1\% & 0.061 \\
Historical best & Coding & 28.6\% & 0.232 & 46.7\% & 0.127 & 60.0\% & 0.057 \\
OrchBench & Research & 64.3\% & 0.062 & 73.3\% & 0.036 & 93.3\% & 0.004 \\
Historical best & Research & 40.0\% & 0.080 & 80.0\% & 0.022 & 93.3\% & 0.009 \\
\bottomrule
\end{tabular}
}
\end{table*}

\begin{table}[t]
\centering
\small

\begin{subtable}[t]{0.72\linewidth}
\centering
\caption{Predictors}
\label{tab:difficulty-predictors}
\begin{tabular}{lccc}
\toprule
Predictor & MAE $\downarrow$ & Pearson $\uparrow$ & Spearman $\uparrow$ \\
\midrule
Scenario mean & 0.088 & 0.725 & 0.614 \\
Complexity linear & 0.088 & 0.726 & 0.691 \\
\bottomrule
\end{tabular}
\end{subtable}
\begin{subtable}[t]{0.72\linewidth}
\centering
\caption{Feature correlations}
\label{tab:difficulty-features}
\begin{tabular}{lcc}
\toprule
Feature & Pearson & Spearman \\
\midrule
Mean sim. makespan & -0.754 & -0.832 \\
Complexity index & -0.761 & -0.823 \\
Tool calls & -0.754 & -0.819 \\
Tool-task ratio & -0.758 & -0.819 \\
Total tokens & -0.729 & -0.773 \\
Task count & -0.717 & -0.701 \\
\bottomrule
\end{tabular}
\end{subtable}
\caption{Task difficulty estimation from lightweight benchmark signals.}
\label{tab:difficulty-estimation}
\end{table}

\section{C. Simulation-Guided Applications}
\label{sec:Simulation_application}

\subsection{I. Task-conditioned Model Selection}

Task-conditioned model selection asks whether \bench{} can recommend a suitable model for a specific task before running all candidate models in the real environment. 
For each task, \bench{} ranks the six models from the simulated orchestration signals. 
Top-\(K\) coverage measures whether the true best real-run model appears in the recommended top-\(K\) set, and Top-\(K\) regret measures the score gap between the best model in that set and the true best model.

We compare \bench{} with a historical-best baseline, which ranks models by their real scores on all other tasks. 
This baseline is strong because it uses real execution history, but it is also expensive and unavailable for a newly introduced task suite. 
Table~\ref{tab:utility-model-selection} shows that \bench{} is more useful when the goal is to choose one model or a small candidate set. 
Overall Top-1 coverage improves from 38.6\% to 61.5\%, and coding improves from 28.6\% to 50.0\%. 
At Top-3, the historical baseline becomes competitive, suggesting that global model strength is helpful once the candidate set is wide. 
Thus, the main value of \bench{} in this setting is task-conditioned early filtering without requiring prior real executions for the target task.

\subsection{II. Task Difficulty Estimation}

Task difficulty estimation asks whether \bench{} can predict how hard a task is before full real evaluation. 
We define task difficulty by the real average score of observed models on the task: a higher score means an easier task, and a lower score means a harder task. 
For each task, we compute a complexity index by averaging four normalized quantities: task count, total tokens, tool calls, and mean simulated makespan. 
A task is therefore more complex if it has more steps, uses more tokens or tools, or takes longer in simulation. 
We then use leave-one-task-out validation: each time, we train a simple linear predictor on 44 tasks and predict the real average score of the held-out task.

Table~\ref{tab:difficulty-estimation} reports the results. 
The scenario mean is already a useful predictor of task difficulty. 
Adding structural complexity keeps MAE the same and improves rank correlation, with Spearman increasing from 0.614 to 0.691. 
The feature correlations are negative because the target is real score, where higher means easier, while the features measure task complexity. 
Thus, larger makespan, more tool use, more tokens, and more subtasks are associated with lower real scores. 
This suggests that \bench{} can provide useful task-difficulty estimates before running full real evaluation.

\begin{table}[t]
\centering
\small
\begin{tabular}{lcc}
\toprule
Metric & Pearson & Spearman \\
\midrule
Cross-framework time & -0.104 & -0.086 \\
Cross-framework token & 0.071 & 0.600 \\
SWE-mini time-token & 0.341 & 0.086 \\
Claude Code time-token & -0.126 & -0.314 \\
\bottomrule
\end{tabular}
\caption{Framework dependence of time and token usage. Correlations are computed on shared model-level results.}
\label{tab:framework-time-token}
\end{table}

\section{D. Simulation-to-Real Validation}
\label{sec:Simulation_to_Real_app}

\subsection{I. Metric Definition}
\label{sec:Simulation_to_Real_app_Metric}

We compute every metric at the run level and then average it over runs from the same model before comparing models. Declared Agents (DA) measures how many agents the planner intended to use: in simulation, this is the value of \texttt{plan.num\_agents}, while in the real workflow it is obtained by counting the \texttt{agent(...)} calls in the generated script. Started Agents (SA) and Completed Agents (CA) use the number of active agents in simulation, namely the number of distinct agents that receive at least one subtask. Their real counterparts are obtained from the workflow journal by counting the unique agent identifiers that produce a \texttt{started} event and a \texttt{result} event, respectively. Subagent Delegation Tendency (SDT) compares the simulated declared-agent count with the fraction of landed real runs that actually invoke the workflow tool. Parallel Utilization (PU) is the fraction of the available agent capacity used in simulation, \(A_{\mathrm{active}}/A_{\max}\). For real runs, it is computed from the subagent time intervals: we sum the active time of all subagents and divide it by the maximum possible active time, given by the maximum number of overlapping subagents multiplied by the total workflow span. Workflow Depth (WD) captures how many dependent stages the workflow contains. In simulation, it is the length of the longest dependency chain in the task DAG, normalized by the number of tasks. In real workflows, we first infer handoff edges by checking whether keywords from an earlier subagent's output appear in a later subagent's prompt;
and the longest path in this inferred graph is then normalized by the number of subagents. Information-Loss Rate (ILR) measures how often information is not explicitly passed between agents. In simulation, it is the number of implicit reread events divided by the total number of explicit handoffs and implicit rereads. In real workflows, it is one minus the inferred handoff density, i.e., the fraction of possible earlier-to-later subagent pairs for which no handoff is detected. Pearson's \(r\) and Spearman's \(\rho\) are computed over the model-level averages; the former compares their numerical values, whereas the latter compares their rankings. The reported \(p\)-values come from two-sided exact permutation tests over the six models.

\subsection{II. Framework Dependence}
\label{sec:Simulation_to_Real_app_framework}

We also test whether time and token usage are stable across execution frameworks. 
Table~\ref{tab:framework-time-token} reports four diagnostic correlations on the shared set of models. 
Cross-framework time and token measure whether model-level resource usage is consistent between SWE-mini and Claude Code dynamic workflow. 
The two time-token rows measure whether models that use more tokens also take more time within each framework. 
Time is almost uncorrelated across frameworks, and the relation between time and token usage is also inconsistent. 
This suggests that time and token usage are strongly affected by framework-specific execution behavior, so we use final score as the main sim-to-real validation target and treat time and token metrics as diagnostics.

\begin{figure*}[!t]
\centering
\includegraphics[width=0.9\textwidth]{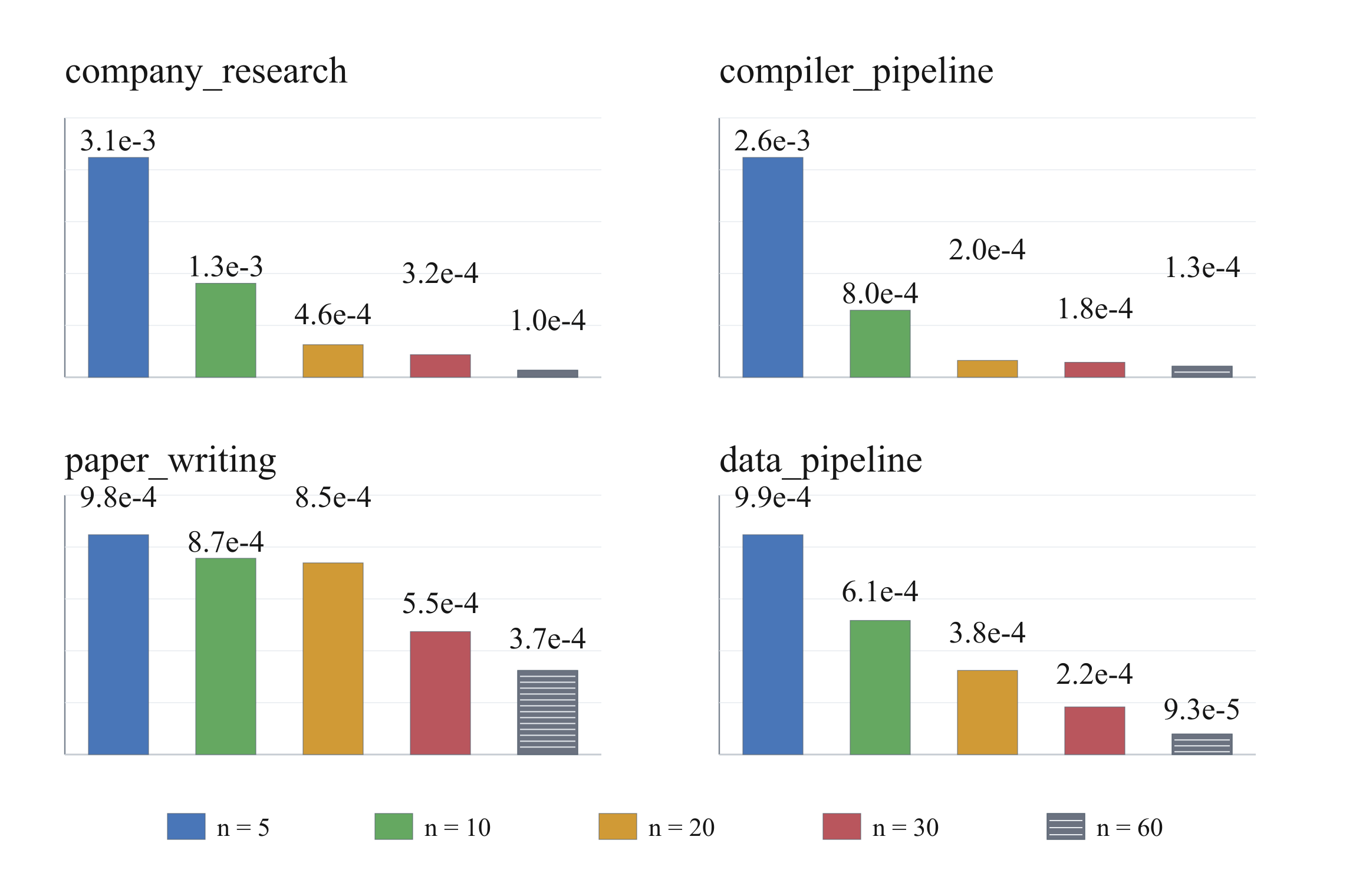}
\caption{Variance of benchmark mean scores under different task counts. }
\label{fig:task-count-stability}
\end{figure*}

\begin{figure*}[t]
  \centering
  \includegraphics[width=0.9\textwidth]{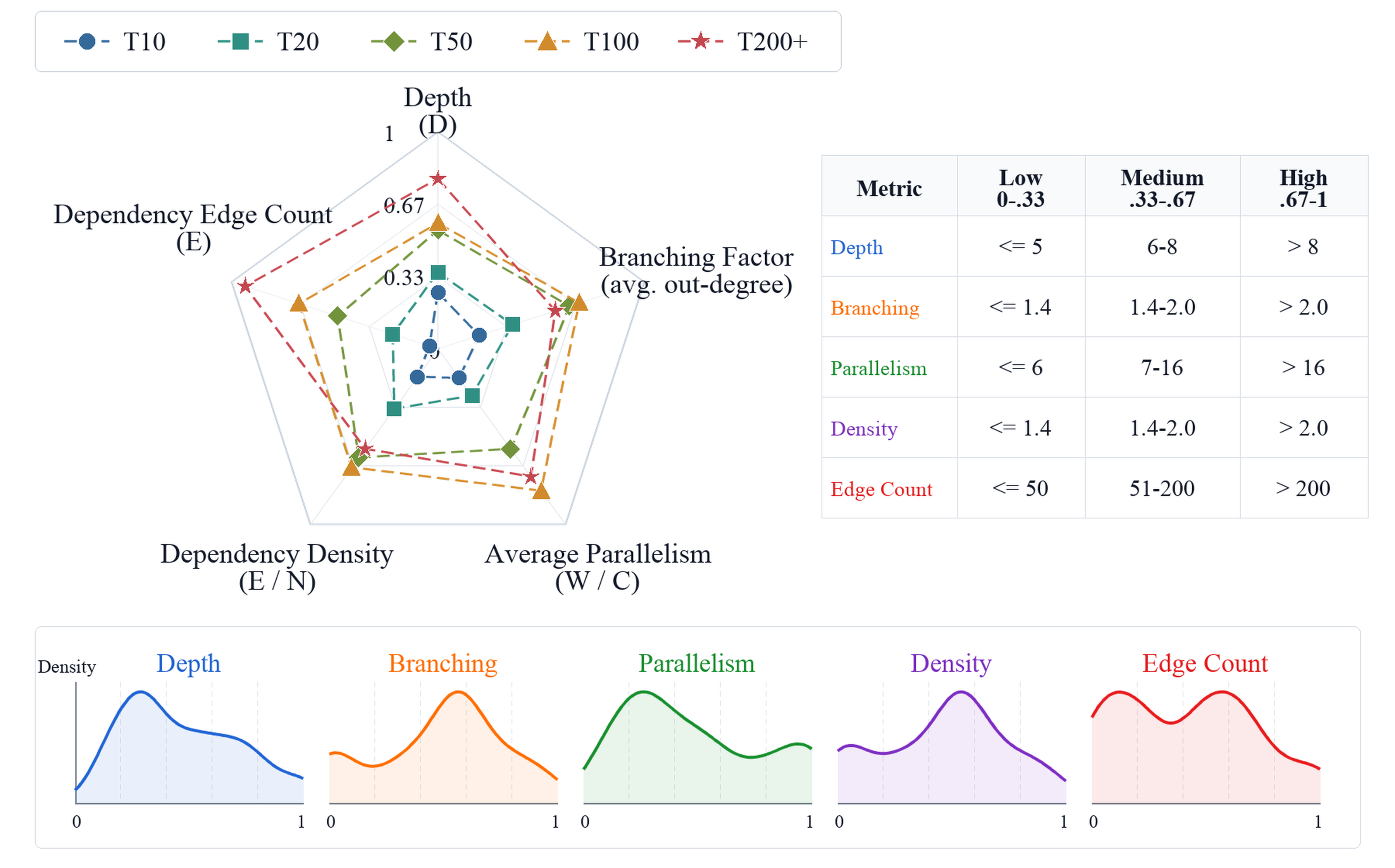}
  \caption{Structural profile of the generated DAG pool. The radar chart reports normalized structural metrics across target task-count groups, and the table gives the three-level reference ranges used for interpretation. The lower panels show the distribution of each normalized metric over the generated DAGs.}
  \label{fig:dag-structural-profile}
\end{figure*}

\begin{algorithm}[t]
\caption{Refinement-based problem generation}
\label{alg:generation}
\small
\begin{algorithmic}[1]
\Statex \textbf{Input:} raw problem pool \(\mathcal{D}\), target subtask count \(n\), target parallelism bucket \(b\), retry budget \(K\)
\Statex \textbf{Output:} accepted problem package \(P=(G,L)\), or no problem
\Statex \textbf{Each generated subtask has:} description, input tokens, execution tokens, result tokens, time cost, and compression class.
\Repeat
    \State sample a raw seed task \(s\) from \(\mathcal{D}\)
    \State ask the seed judge whether \(s\) is suitable for DAG decomposition
\Until{the seed judge accepts \(s\)}
\State ask the generator to build an initial semantic DAG \(G_0\) from \(s\), \(n\), and \(b\)
\For{\(i=0,\ldots,K\)}
    \State run the format judge on \(G_i\)
    \State run the semantic judge on \(G_i\); keep its repair advice
    \If{both judges accept, \(G_i\) has \(n\) subtasks, and \(k_{99}(G_i)/n\in b\)}
        \State set the context limit \(L\) from the generated token budgets
        \State return the accepted problem package \(P=(G_i,L)\)
    \EndIf
    \State ask the generator to revise \(G_i\) using the judge failures and repair advice
    \State call the revised graph \(G_{i+1}\)
\EndFor
\State return no problem if no refined graph passes within \(K\) rounds
\end{algorithmic}
\end{algorithm}

\begin{figure*}[p]
  \centering

  \begin{subfigure}[t]{0.9\textwidth}
    \centering
    \includegraphics[width=\linewidth, trim=0 0 0 115bp, clip]{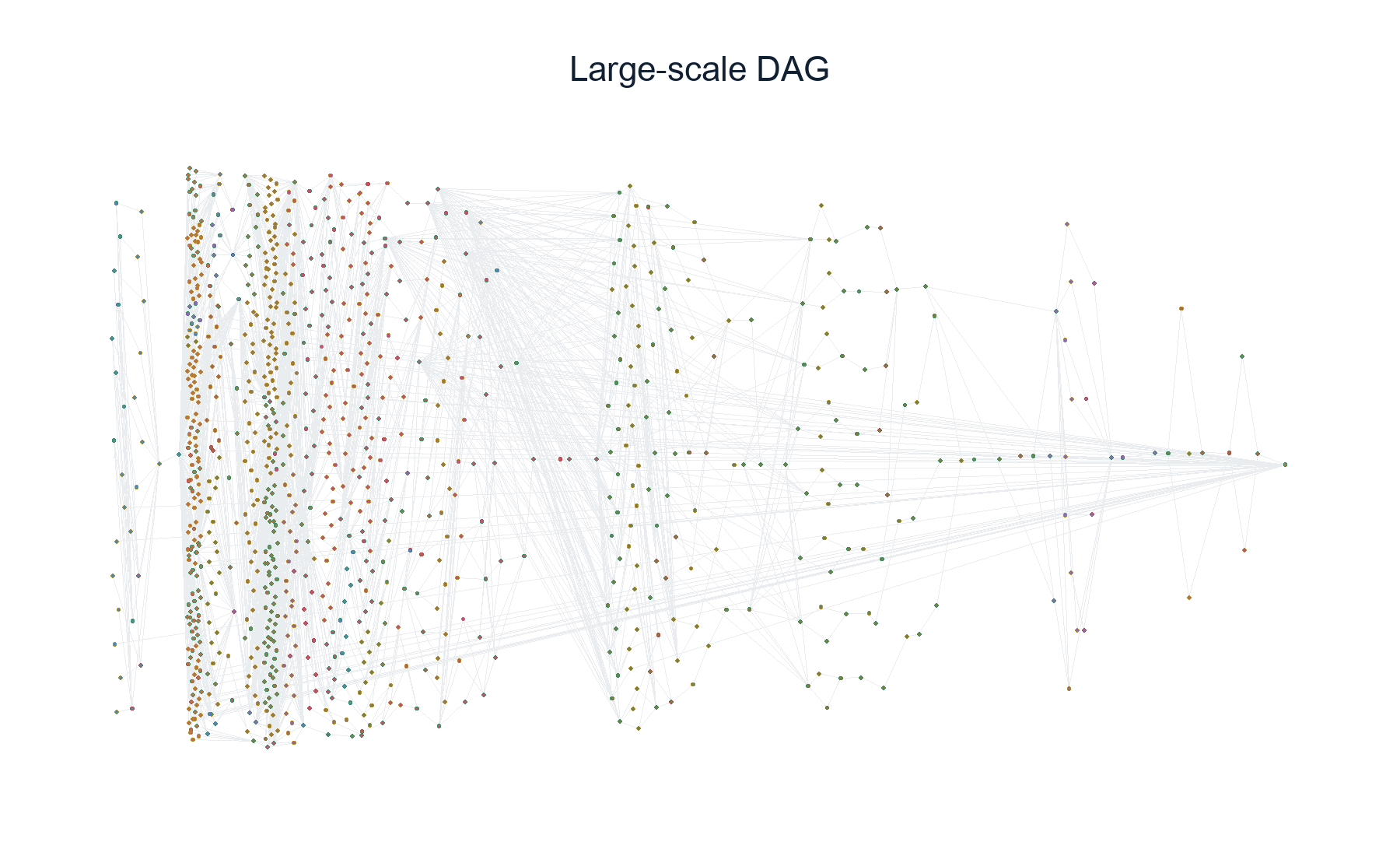}
    \caption{Large-scale DAG.}
  \end{subfigure}

  \begin{subfigure}[t]{0.9\textwidth}
    \centering
    \includegraphics[width=\linewidth, trim=0 0 0 115bp, clip]{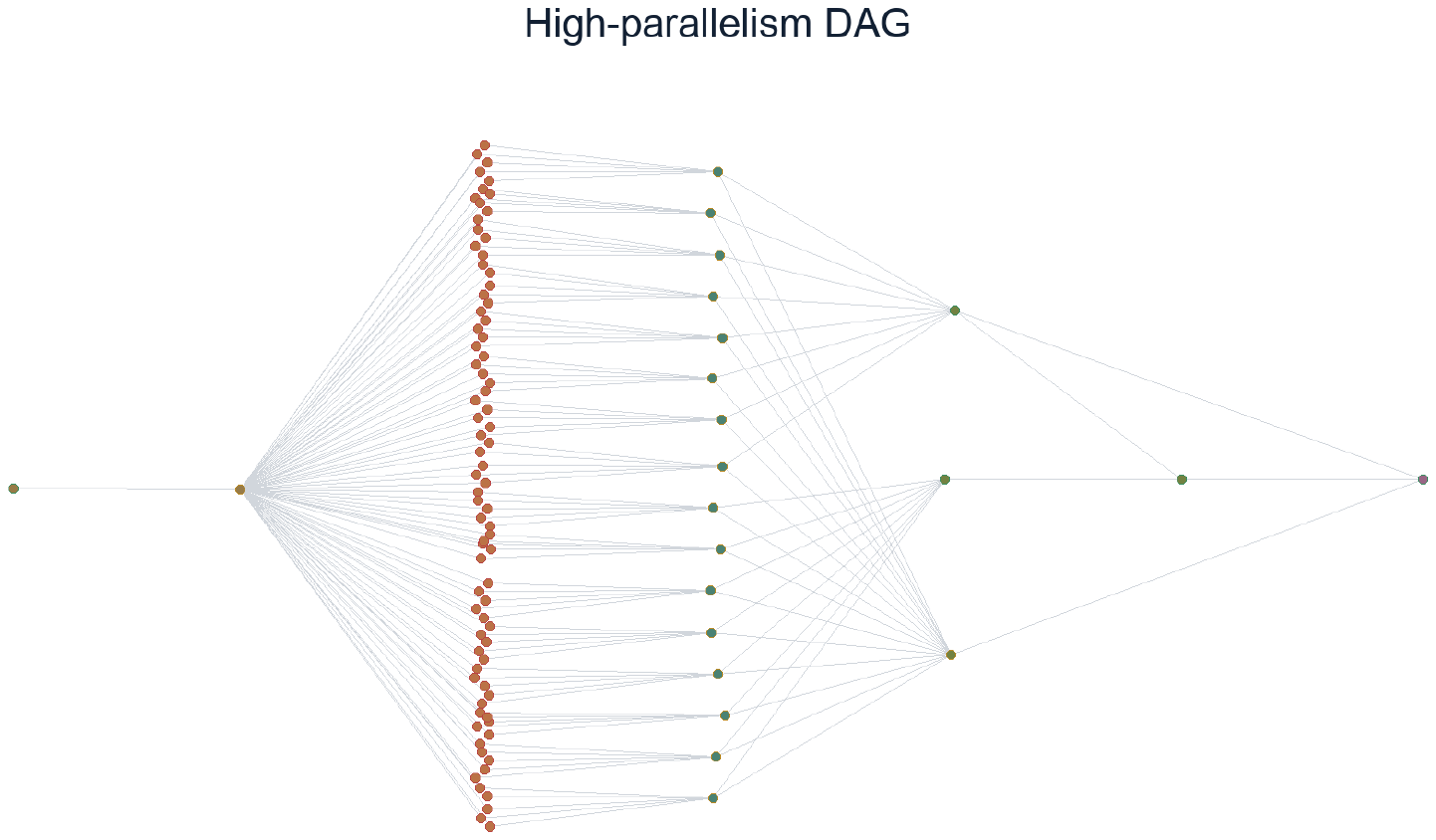}
    \caption{High-parallelism DAG.}
  \end{subfigure}

\end{figure*}

\begin{figure*}[p]
  \ContinuedFloat
  \centering

  \begin{subfigure}[t]{0.9\textwidth}
    \centering
    \includegraphics[width=\linewidth, trim=0 0 0 110bp, clip]{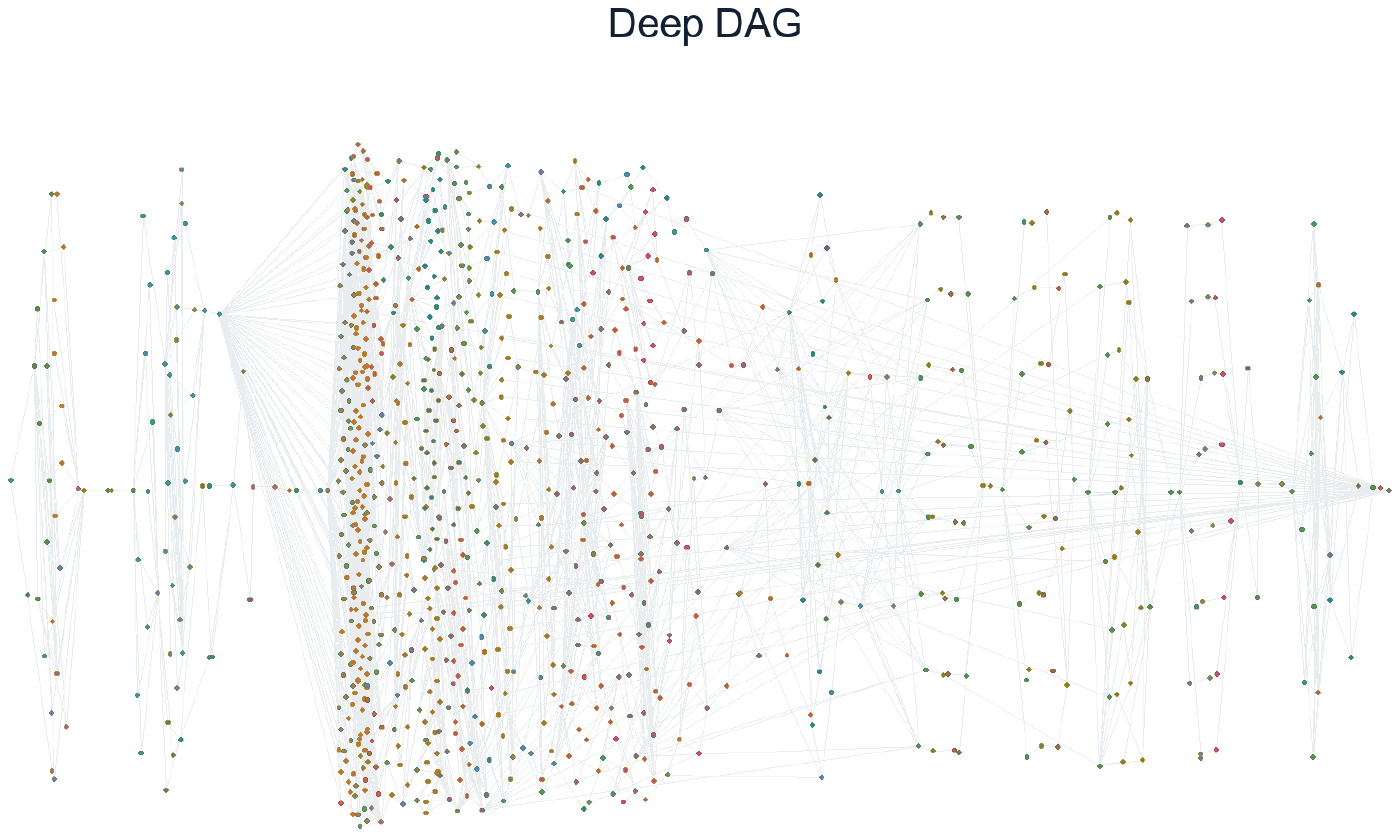}
    \caption{Deep DAG.}
  \end{subfigure}

  \begin{subfigure}[t]{0.9\textwidth}
    \centering
    \includegraphics[width=\linewidth, trim=0 0 0 115bp, clip]{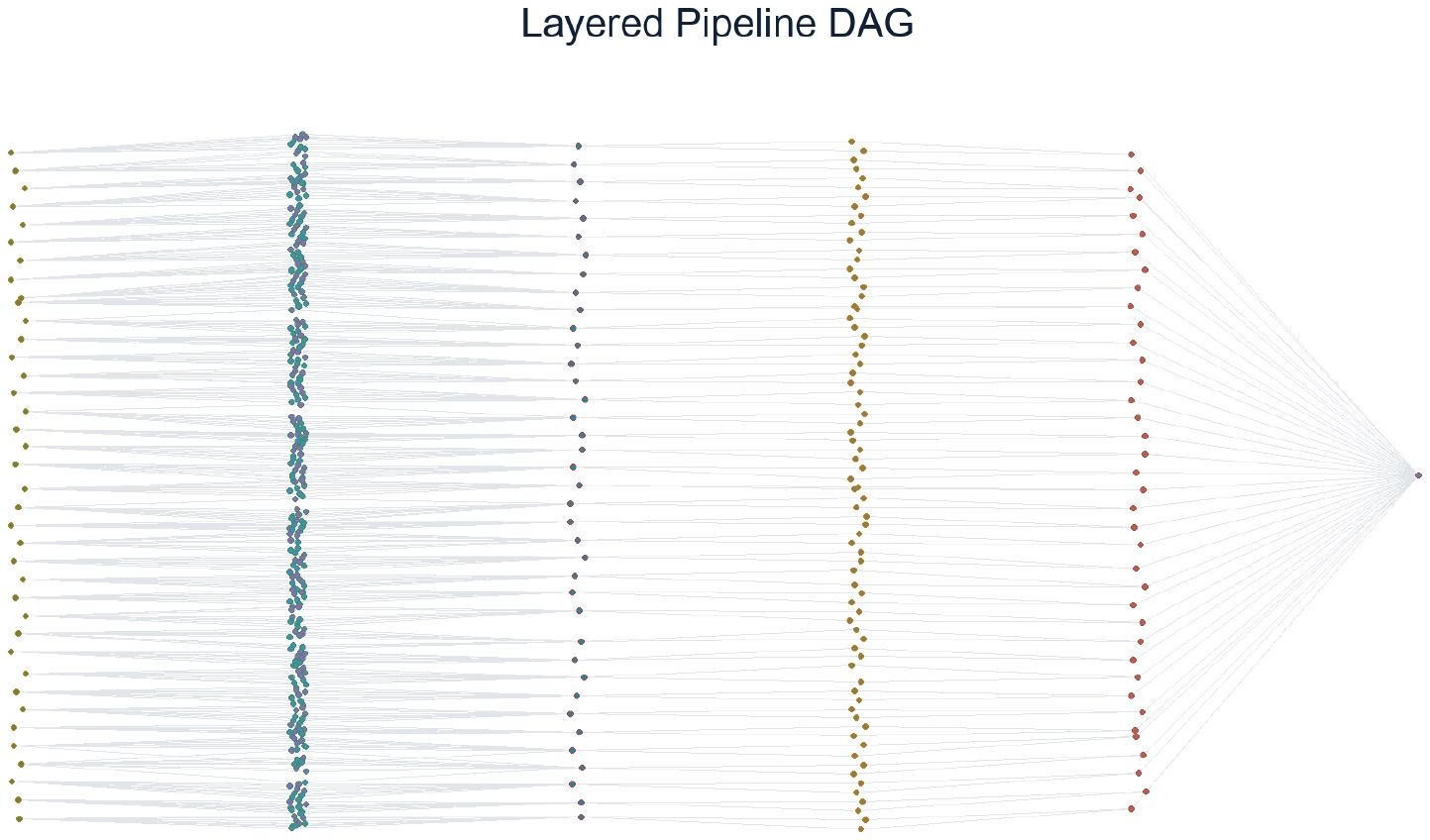}
    \caption{Layered pipeline DAG.}
  \end{subfigure}

\end{figure*}

\begin{figure*}[!t]
  \ContinuedFloat
  \centering

  \begin{subfigure}[t]{0.9\textwidth}
    \centering
    \includegraphics[width=\linewidth, trim=0 0 0 115bp, clip]{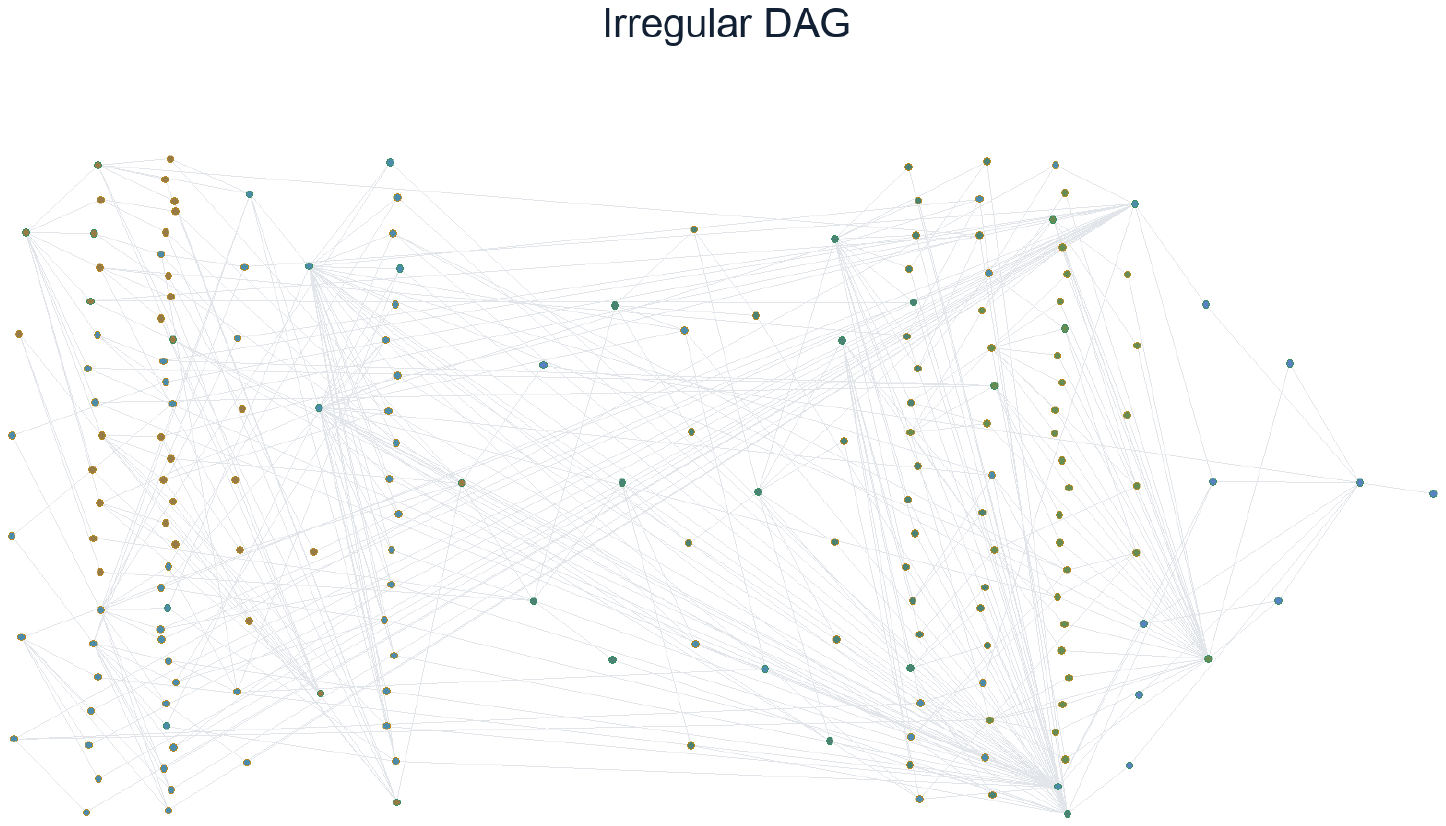}
    \caption{Irregular DAG.}
  \end{subfigure}

  \caption[]{Representative generated DAG structures, continued. The examples illustrate large scale, high parallelism, long dependency depth, regular layered execution, and irregular cross-stage dependency patterns.}
  \label{fig:dag-representative-examples}
\end{figure*}

\section{E. Benchmark Construction}
\label{sec:DAG_construction_app}

\subsection{I. Overview}
\label{sec:DAG_construction_app_overview}

\subsubsection{Task Count Stability}
\label{sec:task-count-stability}

We also check whether the current number of tasks per scenario is sufficient to give a stable benchmark estimate. 
For each scenario, we fix the 60 tasks scored by \texttt{Deepseek-V4-Flash} and use each task's final score as one observation. 
In each repeat, we sample 60 tasks with replacement from this pool. 
We then take the first \(n\in\{5,10,20,30,60\}\) tasks from the sampled sequence and compute their mean final score. 
This process is repeated 10 times, giving 10 benchmark means for each scenario and each task count. 
We report the unbiased sample variance of these 10 means.
Figure~\ref{fig:task-count-stability} shows that the variance generally decreases as the number of tasks increases. 
The reduction is clear in all four scenarios, although paper writing decreases more slowly than the other categories. 
At the full 60-task setting, the estimated variance is small in every scenario, with the largest value being \(3.75\times 10^{-4}\). 
This suggests that 60 tasks per scenario provide a stable estimate of model performance for our benchmark.

\begin{table*}[!t]
\centering
\footnotesize
\begin{tabularx}{\textwidth}{
p{0.15\linewidth}
p{0.08\linewidth}
p{0.22\linewidth}
>{\raggedright\arraybackslash}X
>{\raggedright\arraybackslash}X}
\toprule
Failure mode & Frequency & Original task & Evidence & Interpretation \\
\midrule
Dropped branches before final synthesis
& 5.0\%
& Identify the French foreign name of a black-border creature card illustrated by Matthew D. Wilson.
& The LLM judge reports that several group-aggregation branches never reach the final analysis or delivery node.
& The DAG is executable, but part of the searched information is silently lost before the final answer. \\

\midrule
Artificial over-decomposition
& 6.3\%
& Identify the district with the second-highest number of crimes in 1995 and count male clients in that district.
& A small SQL-style task is expanded into many near-identical query, counting, and correlation subtasks.
& The graph reaches the target size, but the decomposition is not natural for the original task. \\

\midrule
Redundant verification or repeated templates
& 9.6\%
& Compile all singers, episodes, and songs performed in the Chinese TV show \textit{I Am a Singer}/\textit{Singer} from 2013 to 2024.
& The DAG contains a large verification layer that nearly mirrors the extraction layer.
& Verification is useful, but duplicating most extraction tasks adds size without adding much orchestration structure. \\

\midrule
Arbitrary split not grounded in the seed
& 6.7\%
& Apply \texttt{MinMaxScaler} to columns \texttt{A2} and \texttt{A3} grouped by month in a pandas DataFrame.
& The DAG introduces specific months that are not required by the original task.
& The graph looks detailed, but some details are not grounded in the seed task. \\

\midrule
Artificial global synchronization
& 5.0\%
& Complete a large multi-domain web research task with several independent research requests.
& The LLM judge identifies a global bottleneck where independent domains must wait for a shared synchronization task.
& Large DAGs may become wide but still unnatural if they add unnecessary barriers between independent branches. \\

\midrule
Too-fine shallow subtasks
& 1.7\%
& Produce a detailed research paper on supervised topic models for classification and regression from crowds.
& The granularity diagnostic flags many short literature-review nodes as too small or underspecified.
& The task boundaries become dominated by bookkeeping rather than meaningful subtasks. \\

\bottomrule
\end{tabularx}
\caption{Representative DAG generation failure modes.}
\label{tab:dag-generation-failure-cases}
\end{table*}

\begin{table*}[t]
\centering
\footnotesize
\begin{tabularx}{\textwidth}{
p{0.08\linewidth}
p{0.24\linewidth}
>{\raggedright\arraybackslash}X
>{\raggedright\arraybackslash}X}
\toprule
Model & Original task & Evidence & Failure pattern \\
\midrule
Claude Opus 4.8
& Produce a detailed research paper on supervised topic models for classification and regression from crowds.
& The model uses 62 active agents for a low-parallelism problem. It has no missing transfers, but triggers 93 compression events and 97 warnings. Quality drops to 0.206 and final score is 0.409.
& Claude usually writes complete transfers, but can over-fragment low-parallelism writing tasks. The resulting context pressure causes heavy compression and lowers quality. \\

\midrule
DeepSeek-V4-Flash
& Analyze Airbnb's gross booking value per room night from FY2022 to FY2024.
& The model uses 75 active agents for the DAG with a target natural parallelism of 33 and declares 135 transfers, but the simulator records 60 implicit reads and 73 warnings. Quality is 0.511.
& The plan exploits parallelism, but misses many required dependency transfers. The simulator must fall back to penalized reads, so quality is lower than the speed score suggests. \\

\midrule
Kimi 2.6
& Analyze Airbnb's gross booking value per room night from FY2022 to FY2024.
& The model uses 61 active agents and declares 135 transfers, but still has 57 implicit reads. Quality is 0.530 and final score is 0.619.
& Kimi is stronger than most models overall, but this case shows the same transfer-coverage weakness: assigning dependent subtasks to different agents is not enough unless all required information is explicitly passed. \\


\midrule
Doubao Mini
& Analyze Airbnb's gross booking value per room night from FY2022 to FY2024.
& The model uses 85 active agents for the DAG with a target natural parallelism of 33, declares only 123 transfers, and produces 73 implicit reads. Quality is 0.329.
& Doubao tends to over-launch agents while under-specifying communication. This creates high coordination overhead and severe information loss. \\

\bottomrule
\end{tabularx}
\caption{Representative orchestration failures in simulation.}
\label{tab:simulation-failure-cases}
\end{table*}

\subsubsection{Data profile.}
Figure~\ref{fig:dag-structural-profile} profiles the generated DAG pool along five structural metrics. 
The choice of these metrics follows the way recent agent benchmarks and workflow-generation studies frame multi-agent difficulty. MultiAgentBench (ACL 2025) emphasizes that multi-agent evaluation should capture coordination and interaction structure, not only final task completion. WorFBench / Benchmarking Agentic Workflow Generation (ICLR 2025) is especially relevant because it treats agentic workflows as graph-structured subtask dependencies and evaluates workflow quality at both chain and graph levels. TheAgentCompany (NeurIPS 2025 Datasets and Benchmarks) and OdysseyBench (2025) similarly stress realistic, long-horizon work settings where agents must coordinate across tools, subtasks, and dependencies. Motivated by this view, we characterize each generated problem as a DAG and measure the structural properties that directly affect orchestration difficulty.

Specifically, depth $D$ measures the number of topological dependency layers and captures the sequential horizon of a problem. Branching factor is the average out-degree over non-terminal tasks, reflecting how broadly a problem decomposes into downstream subtasks. Average parallelism $W/C$ measures total work $W$ divided by critical-path length $C$, which is the standard work-over-critical-path measure used in DAG scheduling to estimate exploitable concurrency. Dependency density $E/N$ normalizes the number of dependency edges by the number of tasks and measures average coupling strength. Edge count $E$ measures the total number of precedence constraints and captures the overall dependency volume. Together, these metrics cover the main structural sources of orchestration difficulty: long dependency chains, broad decomposition, parallel execution opportunities, and dense cross-task constraints.

For readability, we convert each metric into three reference regimes: Low, Medium, and High. 
This three-level discretization is used only for interpretation and normalization in the figure; it is not used to rebalance the dataset or force a uniform distribution. The thresholds are simple structural reference points: for example, $D>8$ already indicates a long sequential horizon, $W/C>16$ indicates substantial exploitable parallelism, and $E/N>2.0$ indicates a strongly coupled dependency graph. 
Values above the High threshold are clipped to the outer ring of the radar chart, so reaching the boundary should be interpreted as entering the high-complexity regime rather than matching the single largest DAG.

The labels T10, T20, T50, T100, and T200+ denote target task-count groups used during generation. 
T10 contains DAGs generated around the 10-task scale, T20 around the 20-task scale, and similarly for T50 and T100. T200+ aggregates larger target scales starting from 200 tasks, since these large-scale settings are generated less densely but are important for showing the upper structural range. The radar curves show how structural difficulty changes across task scales, while the lower distribution panels show the overall spread of normalized metric values across the generated DAG pool. The figure therefore demonstrates that the dataset is not concentrated in a single graph pattern, but covers diverse levels of depth, branching, dependency volume, coupling strength, and parallel execution potential.

Figure~\ref{fig:dag-representative-examples} complements the aggregate statistics with representative DAG layouts drawn from the original generated problem graphs. 
The selected examples make the structural range visually explicit: the large-scale DAG demonstrates thousand-node orchestration, the high-parallelism DAG exposes many concurrently schedulable branches, the deep DAG highlights long dependency chains and multi-step progression, the layered pipeline DAG shows regular batched fan-out and fan-in, and the irregular DAG illustrates heterogeneous cross-stage dependencies. 
Together, these examples reinforce the aggregate profile by showing that the generated problems cover diverse graph sizes, dependency depths, structural patterns, and parallel execution opportunities.

\begin{table*}[t]
\centering
\scriptsize
\resizebox{0.9\textwidth}{!}{
\begin{tabular}{llcccccc}
\toprule
Target size & Judge & Overall & Coverage & Missing & Dependency & Redundancy & Parallelism \\
\midrule
10   & Rule-based & 83.00 & 97.00  & 98.00 & 94.00 & 92.00 & 96.00 \\
10   & Gemini     & 98.40 & 99.30  & 99.50 & 99.30 & 96.50 & 98.00 \\
10   & GLM        & 84.50 & 88.10  & 86.80 & 87.70 & 80.90 & 85.40 \\
\midrule
20   & Rule-based & 63.81 & 99.02  & 99.50 & 75.43 & 89.36 & 100.00 \\
20   & Gemini     & 95.12 & 97.00  & 98.50 & 96.38 & 90.00 & 96.88 \\
20   & GLM        & 77.44 & 87.22  & 88.33 & 73.11 & 67.22 & 83.00 \\
\midrule
50   & Rule-based & 77.69 & 98.78  & 99.80 & 97.90 & 81.01 & 97.60 \\
50   & Gemini     & 97.33 & 99.00  & 99.67 & 98.83 & 92.83 & 98.50 \\
50   & GLM        & 82.50 & 89.00  & 85.20 & 86.70 & 69.20 & 86.00 \\
\midrule
100  & Rule-based & 63.06 & 98.68  & 99.79 & 87.30 & 75.71 & 87.40 \\
100  & Gemini     & 89.67 & 97.83  & 97.83 & 93.33 & 83.83 & 88.83 \\
100  & GLM        & 76.11 & 88.00  & 90.11 & 78.33 & 62.33 & 80.00 \\
\midrule
200  & Rule-based & 86.15 & 100.00 & 100.00 & 93.53 & 92.25 & 99.42 \\
200  & Gemini     & 95.75 & 97.88  & 98.12 & 94.50 & 93.88 & 96.75 \\
200  & GLM        & 85.00 & 90.78  & 89.22 & 83.67 & 79.44 & 89.67 \\
\midrule
500  & Rule-based & 67.92 & 90.12  & 99.76 & 75.39 & 75.12 & 67.14 \\
500  & Gemini     & 71.50 & 95.00  & 97.50 & 97.50 & 57.50 & 75.00 \\
500  & GLM        & 75.75 & 85.50  & 78.75 & 87.50 & 64.00 & 74.50 \\
\midrule
1000 & Rule-based & 85.20 & 100.00 & 100.00 & 79.33 & 77.89 & 99.13 \\
1000 & Gemini     & 93.00 & 98.00  & 95.00 & 92.00 & 85.00 & 96.00 \\
1000 & GLM        & 80.60 & 89.60  & 91.60 & 76.70 & 69.40 & 88.50 \\
\bottomrule
\end{tabular}
}
\caption{Validation results for generated DAGs. Higher is better.}
\label{tab:dag-validation}
\end{table*}

\begin{table}[t]
\centering
\small
\resizebox{0.5\textwidth}{!}{
\begin{tabular}{lcccc}
\toprule
Setting & Accuracy & Average & $\Delta$ Acc & $\Delta$ Score \\
\midrule
w/o DAG   & 2/10 (20.00\%) & 3.300 & -- & -- \\
w/ DAG & 3/10 (30.00\%) & 3.575 & +10.00 pp & +0.275 \\
\bottomrule
\end{tabular}}
\caption{Effect of providing DAGs in actual agent runs.}
\label{tab:dag-real-run}
\end{table}

\subsection{II. DAG Generation}
\label{sec:DAG_construction_app_generation}

Algorithm~\ref{alg:generation} describes how \bench{} builds DAG problems from real seed tasks. 
We first sample a seed task \(s\) from the problem pool \(\mathcal{D}\). 
A seed judge filters out tasks that are not suitable for DAG generation. 
For each accepted seed, the generator produces a DAG with a target number of subtasks \(n\) and a target level of natural parallelism. 
Natural parallelism measures how many agents a DAG can use effectively. 
We define it as \(k_{99}/n\). 
Here \(k_{99}\) is the smallest number of workers whose schedule gets within one percent of the critical-path lower bound:
\[
k_{99}(G)=\min\{k:\mathrm{ms}_k(G)\le C(G)+0.01(W(G)-C(G))\}.
\]
In this equation, \(W(G)\) is the sum of all task execution times, \(C(G)\) is the critical-path time, and \(\mathrm{ms}_k(G)\) is the makespan of a deterministic list schedule with \(k\) workers. 
A larger \(k_{99}/n\) means the graph has more useful parallelism.

Generation then proceeds as a refinement loop. 
The structural judge checks whether the DAG is well formed. 
The semantic judge checks whether the DAG is a meaningful decomposition of the seed task and whether its natural parallelism matches the target. 
The generator revises the DAG using judge feedback until both judges accept it. 
The final package contains the accepted DAG and its context limit \(L\), which are passed to the simulation stage.

\subsection{III. Generation Failure Cases}
We analyze failure cases from two sides.
The first is DAG generation: whether the generated graph is a meaningful decomposition of the original task.
The second is orchestration simulation: whether the tested model writes a plan that uses agents, transfers, and compression properly.
These failures help explain both the limitations of the generated benchmark problems and the common weaknesses of current models when they are asked to orchestrate multiple agents.

\subsubsection{DAG Generation Failures}
\label{app:dag-generation-failures}

Most generated DAGs pass basic structural checks, such as schema validity, acyclicity, and terminal reachability.
The remaining failures are mainly semantic.
They happen when the graph reaches the target size but does so in a way that is not fully faithful, natural, or useful for orchestration evaluation.
The failure modes in Table~\ref{tab:dag-generation-failure-cases} are not mutually exclusive: a DAG may contain several issues at the same time.

To avoid making overlapping failure modes look like independent failure rates, we report normalized frequencies.
We first compute the DAG-level failure rate over the 70 sampled DAGs: 24 DAGs are rejected by at least one successful LLM judge, giving \(p_{\mathrm{fail}}=24/70=34.3\%\).
We then count how often each failure mode appears among the failed-case annotations, giving \(C=82\) total error mentions.
For a failure mode with count \(c_m\), the reported frequency is \(p_{\mathrm{fail}}\cdot c_m/C\).
Thus the frequencies in Table~\ref{tab:dag-generation-failure-cases} decompose the overall failed-case mass and sum to \(34.3\%\), rather than exceeding 100\% because of co-occurring errors.

These cases show that DAG generation is less likely to fail at the level of graph syntax, and more likely to fail at the level of task semantics.
In particular, large DAGs tend to expose more redundancy and less clean dependency structure, while smaller DAGs are more likely to preserve the original task boundary.

\subsubsection{Orchestration Simulation Failures}
\label{app:simulation-failure-cases}

We also inspect workflow plans written by tested models.
Here the DAG is fixed, so failures mainly come from the model's orchestration decisions.
Common problems include launching too many agents, splitting dependent subtasks across agents without declaring transfers, using compression too aggressively, and creating plans that look parallel but lose important upstream information.
Table~\ref{tab:simulation-failure-cases} gives representative cases.

\subsection{IV. DAG Validation}
\label{sec:DAG_construction_app_validation}

The generated DAGs are the basis of \bench{}, since every orchestration plan is evaluated on top of their dependency structure. 
It is necessary to check whether these DAGs are structurally valid and semantically reasonable. 
We evaluate them along six dimensions: structure; coverage; missing-subtasks; dependency correctness; redundancy; and parallelism plausibility. 
The six validation dimensions are evaluated using both deterministic rules and LLM-based judging. 
For the rule-based evaluation, each generated DAG is scored independently. 
Let \(N\) be the actual number of subtasks in the DAG. 
For each dimension, the checker counts the affected task nodes and additional structural violations, and converts them into a percentage using \(100\times(1-\text{issues}/N)\), with the number of issues capped at \(N\). 
The values reported in Table~\ref{tab:dag-validation} are the average percentages over the sampled DAGs at each target size. 
Thus, these values are not pass rates. 
The \textit{Overall} score is also not the arithmetic mean of the other columns: it combines the task-level issues and structural penalties from all six dimensions, including structure, coverage, missing subtasks, dependency correctness, redundancy, and parallelism plausibility. 
Structure is included in the overall score but is not shown as a separate column in the table.

The structure dimension checks whether the generated object is a valid DAG with a reasonable size and complete task metadata. 
The task count is considered valid when it falls within \(10\%\) of the target count, with at least one task of tolerance. 
The checker then verifies that task identifiers are unique, every dependency refers to an existing task, no task depends on itself, the graph is acyclic, there is exactly one terminal task, and every task can reach that terminal task. 
It also checks that input, execution, and result token counts and time costs are positive, and that compression and tool fields use valid values. 
Any task involved in one of these violations contributes to the structure penalty; graph-level violations such as a cycle, an incorrect terminal count, or a task-count mismatch are added as additional penalties.

Coverage measures whether the generated subtasks preserve the content of the original task. 
The rule-based checker extracts the 40 most frequent non-stopword tokens from the original seed prompt, dataset name, and summary, and computes the fraction that appear in the generated title, goal, stages, descriptions, or tool fields. 
A keyword coverage of at least \(0.35\) is required. 
The checker also assigns each subtask coarse functional roles, such as acquisition, analysis, validation, synthesis, and delivery, and verifies that all category-specific required roles are present. 
The coverage score penalizes the gap below the \(0.35\) keyword threshold and each missing required role.

Missing-subtasks measures whether the DAG omits an essential part of the workflow. 
The rule-based proxy flags tasks that do not lead to the terminal task, missing category-specific roles, and graphs with an incorrect terminal structure. 
Therefore, this score mainly captures missing workflow stages and disconnected parts of the graph; it is not based on counting every subtask that could have been generated.

Dependency correctness evaluates whether the edges form a plausible prerequisite structure. 
The graph must contain at least \(N-1\) edges, which is the minimum required to connect \(N\) tasks into a tree-like dependency structure. 
For each edge, the checker infers a coarse role order from the task descriptions, using acquisition, analysis, validation, synthesis, and delivery as progressively later stages. 
An edge is treated as a backward edge when its parent appears more than one role stage later than its child. 
The affected endpoints of such edges are penalized, and the backward-edge rate must not exceed \(0.08\). 
The checker also limits extreme fan-in and fan-out: the proxy limits are \(\max(10,\lceil0.45N\rceil)\) incoming edges and \(\max(10,\lceil0.90N\rceil)\) outgoing edges for a task.

Redundancy measures whether different subtasks repeat essentially the same work. 
Descriptions are normalized before comparison. 
The checker counts exact duplicate descriptions and near-duplicate pairs, where a pair is considered near-duplicate when its normalized text similarity reaches \(0.97\) after token-overlap checks. 
Descriptions referring to different explicit objects, such as different years, quarters, or partitions, are not treated as duplicates. 
The near-duplicate pair ratio is the number of near-duplicate pairs divided by the number of comparable pairs, and it must not exceed \(0.02\). 
The checker additionally forms a structural signature from the stage, tool type, dependency list, and the beginning of the description; repeated signatures are also penalized.

Parallelism plausibility measures whether the DAG exposes useful and coherent parallel work. 
The checker uses deterministic list scheduling based on the task time costs to compute the critical path, total work, layer widths, and \(k_{99}\), the smallest number of agents whose estimated makespan reaches \(99\%\) of the attainable scheduling gain. 
The resulting \(k_{99}\) must fall inside the target parallelism range, which is derived from the declared parallelism bucket or target agent count. 
The DAG must also contain a downstream merge node when its target shape requires parallel branches, and the widest layer must not contain excessive near-duplicate tasks; for this last check, the pair-ratio threshold is \(0.08\).

The Gemini and GLM rows use the same metric names but are produced by LLM-as-judge rather than by these deterministic rules. 
The judge assigns each dimension an integer quality score from 0 to 100, where higher values indicate better coverage, more plausible dependencies, less redundancy, fewer missing stages, and more reasonable parallelism. 
The \textit{Overall} judge score is a holistic quality assessment rather than a deterministic average of the five individual scores. 
The exact judge prompt, DAG representation supplied to the judge, and response normalization procedure are described in the dedicated judge-prompt subsection.

Table~\ref{tab:dag-validation} summarizes the results. 
The scores tend to fluctuate up and down and vary across sizes.
The main failure comes from redundancy and dependency quality, while coverage and missing-subtask scores remain high in most settings. 
This suggests that the generated DAGs usually preserve the main task content, although larger DAGs contain more overlap and less clean dependency structure.
We further test whether the generated DAGs are useful in actual agent runs. 
We compare two settings on 10 tasks: one gives the model only the original task description, and the other gives the generated 500-subtask DAG. 
In both settings, DeepSeek-V4-Pro runs under Claude Code dynamic workflow mode. 
The DAG setting improves accuracy from 20.00\% to 30.00\% and slightly increases the average score, as shown in Table~\ref{tab:dag-real-run}.

\begin{table*}[t]
\centering
\small
\resizebox{0.8\linewidth}{!}{
\begin{tabular}{lcccccc}
\toprule
Bucket & Active agents & Transfer coverage & Missing transfer & Quality & Speed & Final score \\
\midrule
Low    & 39.25 & 0.858 & 23.47 & 0.568 & 0.509 & 0.585 \\
Medium & 62.53 & 0.973 & 6.26  & 0.584 & 0.353 & 0.509 \\
High   & 79.58 & 0.992 & 1.77  & 0.540 & 0.336 & 0.494 \\
\bottomrule
\end{tabular}
}
\caption{100-task results by natural-parallelism bucket.}
\label{tab:topology-regime-100}
\end{table*}

\begin{table}[t]
\centering
\scriptsize
\setlength{\tabcolsep}{3pt}
\renewcommand{\arraystretch}{0.96}
\resizebox{\columnwidth}{!}{%
\begin{tabular}{@{}llrrrr@{}}
\toprule
Context & System & Quality & Speed & Token & Final Score \\
\midrule
\(16\mathrm{k}\)
& Single-agent         & 0.423 & 0.181 & 1.000 & 0.535 \\
& DeepSeek-V4-Flash    & 0.799 & 0.562 & 0.668 & 0.676 \\
& DeepSeek-V4-Pro      & 0.764 & 0.550 & 0.670 & 0.661 \\
& Doubao-Seed-2.0-Mini & 0.651 & 0.533 & 0.684 & 0.623 \\
& GLM-5.1              & 0.790 & 0.547 & 0.671 & 0.669 \\
& Kimi-K2.6            & 0.721 & 0.526 & 0.691 & 0.646 \\
& Qwen3.6-35B-A3B      & 0.624 & 0.546 & 0.706 & 0.626 \\
& Multi-agent average  & 0.725 & 0.544 & 0.682 & 0.650 \\
\midrule
\(32\mathrm{k}\)
& Single-agent         & 0.649 & 0.190 & 1.000 & 0.613 \\
& DeepSeek-V4-Flash    & 0.883 & 0.562 & 0.669 & 0.705 \\
& DeepSeek-V4-Pro      & 0.861 & 0.579 & 0.664 & 0.701 \\
& Doubao-Seed-2.0-Mini & 0.699 & 0.552 & 0.682 & 0.644 \\
& GLM-5.1              & 0.908 & 0.563 & 0.657 & 0.709 \\
& Kimi-K2.6            & 0.803 & 0.539 & 0.681 & 0.674 \\
& Qwen3.6-35B-A3B      & 0.769 & 0.555 & 0.676 & 0.667 \\
& Multi-agent average  & 0.821 & 0.558 & 0.671 & 0.684 \\
\midrule
\(64\mathrm{k}\)
& Single-agent         & 0.792 & 0.193 & 1.000 & 0.662 \\
& DeepSeek-V4-Flash    & 0.918 & 0.581 & 0.661 & 0.720 \\
& DeepSeek-V4-Pro      & 0.904 & 0.568 & 0.652 & 0.708 \\
& Doubao-Seed-2.0-Mini & 0.750 & 0.550 & 0.664 & 0.655 \\
& GLM-5.1              & 0.925 & 0.580 & 0.650 & 0.718 \\
& Kimi-K2.6            & 0.799 & 0.540 & 0.678 & 0.672 \\
& Qwen3.6-35B-A3B      & 0.818 & 0.550 & 0.659 & 0.675 \\
& Multi-agent average  & 0.852 & 0.561 & 0.661 & 0.692 \\
\midrule
\(128\mathrm{k}\)
& Single-agent         & 0.852 & 0.195 & 1.000 & 0.682 \\
& DeepSeek-V4-Flash    & 0.893 & 0.581 & 0.659 & 0.711 \\
& DeepSeek-V4-Pro      & 0.906 & 0.554 & 0.655 & 0.705 \\
& Doubao-Seed-2.0-Mini & 0.784 & 0.553 & 0.667 & 0.668 \\
& GLM-5.1              & 0.927 & 0.563 & 0.652 & 0.714 \\
& Kimi-K2.6            & 0.834 & 0.548 & 0.671 & 0.684 \\
& Qwen3.6-35B-A3B      & 0.813 & 0.547 & 0.676 & 0.678 \\
& Multi-agent average  & 0.859 & 0.558 & 0.663 & 0.693 \\
\bottomrule
\end{tabular}%
}
\caption{Quality, speed, token, and final scores under different context limits. All values are averaged over the 50 sampled problems.}
\label{tab:context-sweep}
\end{table}

\begin{table}[t]
\centering
\caption{Sensitivity to the missing-transfer factor \(\lambda\).
\(Q_{\mathrm{aff}}\) averages the plans containing missing transfers;
\(\sigma_Q\) is computed across all model means, and
\(\Delta Q_{\mathrm{G-D}}\) is the Gemini--Doubao quality gap.}
\label{tab:lambda-sweep}
\small
\begin{tabular}{c@{\quad}ccccc}
\toprule
\(\lambda\) & Quality & Final & \(Q_{\mathrm{aff}}\)
& \(\sigma_Q\) & \(\Delta Q_{\mathrm{G-D}}\) \\
\midrule
0.00 & 0.504 & 0.515 & 0.244 & 0.117 & 0.341 \\
0.25 & 0.527 & 0.522 & 0.311 & 0.105 & 0.317 \\
\textbf{0.50} & \textbf{0.561} & \textbf{0.534}
& \textbf{0.412} & \textbf{0.081} & \textbf{0.247} \\
0.75 & 0.608 & 0.550 & 0.550 & 0.046 & 0.150 \\
1.00 & 0.689 & 0.576 & 0.785 & 0.022 & 0.011 \\
\bottomrule
\end{tabular}
\end{table}

\section{F. Additional Results}
\label{sec:additional}

\subsection{I. Detailed Discussion of the Main Results}

Table~\ref{tab:main-results} reports the main results of \bench{} across target sizes.
Here we provide additional analysis of it.

\paragraph{Transfer coverage matters more than agent count.}
A central finding is that using more agents is not the same as better orchestration. 
Models with similar active-agent counts can have very different quality because their transfer coverage differs. 
For example, on 100-task DAGs, \texttt{gemini-3.1-pro-preview} and \texttt{doubao-seed-2-0-mini} both use about 63 active agents, but their missing-transfer counts are 0.07 and 22.70, and their quality scores are 0.690 and 0.443. 
The difference is not how many agents are used, but whether the information needed by downstream subtasks is passed forward.

\paragraph{Orchestration Depends on Structure and Trade-offs.}
The results also show different orchestration styles. 
\texttt{claude-opus-4.8} is a conservative handoff model: it rarely misses transfers and preserves quality, but does not always achieve the best final score because final score also includes speed and token efficiency. 
By contrast, \texttt{gpt-5.5}, \texttt{glm-5.1}, and \texttt{gemini-3.1-pro-preview} often achieve stronger scores by balancing quality against speed and token cost. 
For example, on 100-task DAGs, \texttt{deepseek-v4-pro} has lower quality than \texttt{claude-opus-4.8}, but a slightly higher final score because its speed and token scores are higher. 
Thus, the final score should be read as a Pareto-style summary, not as a pure quality ranking.

Source structure further changes which model is best. 
No model dominates every source family: \texttt{gpt-5.5} leads on company research, \texttt{gemini-3.1-pro-preview} leads on compiler pipelines and data pipelines, and \texttt{glm-5.1} leads on paper writing. 
The gaps are also small: the top-two gap is only 0.0052 on company research, 0.0014 on compiler pipelines, 0.0007 on data pipelines, and 0.0008 on paper writing. 
This indicates that \bench{} is not measuring a single abstract notion of model strength. 
It is also measuring whether a model can adapt its orchestration plan to different DAG structures and information-flow patterns.

Finally, task count alone does not explain difficulty. 
Table~\ref{tab:topology-regime-100} shows the 100-task results by natural-parallelism bucket. 
Low-parallelism DAGs have the most missing transfers, but still obtain the highest final score because their speed and token scores are better. 
High-parallelism DAGs have almost perfect transfer coverage, but lower final scores because the larger number of agents does not fully compensate for coordination cost. 
This suggests two different failure modes in large DAGs: low-parallelism graphs mainly fail through broken information chains, while high-parallelism graphs mainly expose the cost of coordinating many distributed branches.

Overall, the main result is that better orchestration is not equivalent to launching more agents. 
The stronger models are those that can distribute work while keeping dependency information available to downstream subtasks, and that can choose when the quality gain from additional coordination is worth the speed and token cost. 
This is the behavior that \bench{} is designed to isolate.

\subsection{II. Context-Limit Sweep}
\label{sec:context-limit-sweep}
Table~\ref{tab:context-sweep} shows that the benefit of multi-agent orchestration is conditional on context pressure rather than universal. As the context limit increases from \(16\mathrm{k}\) to \(128\mathrm{k}\), the average multi-agent quality advantage shrinks from \(+0.302\) to \(+0.007\), while its final-score advantage falls from \(+0.116\) to \(+0.011\). Under tight context, distributing intermediate state across agents avoids the repeated compression suffered by a single agent; under large context, this benefit disappears, while multi-agent execution continues to incur communication-token and information-transfer costs. Consequently, at \(128\mathrm{k}\), Doubao, Kimi, and Qwen already have lower average quality than the single-agent baseline, and multi-agent quality is lower in \(82\%\) of all model--problem pairs. The slightly positive overall average is driven by the 100-subtask problems, for which the single agent still undergoes \(240\) compression events on average and multi-agent quality remains \(0.362\) higher. For the 10-, 20-, and 50-subtask problems, multi-agent quality is already lower at \(128\mathrm{k}\). These results indicate that multi-agent orchestration is most useful when it prevents context overflow; once the working state fits within a single context window, the single-agent system generally preserves information more reliably without paying coordination costs.

\subsection{III. Missing-transfer Sweep}

We sweep \(\lambda\in\{0,0.25,0.5,0.75,1\}\) by replaying existing $n=100$ plans, without making additional model calls. 
For each missing cross-agent transfer, the upstream quality contribution is multiplied by \(\lambda\); thus, \(\lambda=0\) applies the strongest penalty, while \(\lambda=1\) removes the quality penalty entirely. 
Since the plans and their speed and token costs remain fixed, the experiment isolates the effect of this parameter.
As shown in Table~\ref{tab:lambda-sweep}, increasing \(\lambda\) mechanically raises Quality and Final but steadily reduces model separation. At \(\lambda=0\) and \(0.25\), the mean quality of affected plans is only \(0.244\) and \(0.311\). Conversely, at \(\lambda=1\), \(\sigma_Q\) falls to \(0.022\), and the Gemini--Doubao quality gap nearly disappears at \(0.011\). With \(\lambda=0.5\), affected-plan quality reaches \(0.412\), while \(\sigma_Q=0.081\) and a quality gap of \(0.247\) are retained. It therefore provides a meaningful penalty without obscuring the broader differences among models, making it a balanced default.

\section{G. Other Settings}
\paragraph{Context and agent budgets.}
Each standard benchmark instance retains the maximum agent budget \(A_{\max}=100\) stored in its problem package. 
The per-agent context limit \(L\) is fixed before orchestration planning and is independent of the evaluated planner. We estimate the active working set from task inputs, retained parent packages, task outputs, and a fraction of accumulated execution history. The raw estimate is the maximum of \(1.35\) times the 90th-percentile active working set, twice the 95th-percentile package size, and an amortized per-agent estimate of cumulative execution and package tokens adjusted by the proportion of compression-fragile tasks. It is then mapped to the smallest accommodating window in
\(\{16\mathrm{k},32\mathrm{k},64\mathrm{k},128\mathrm{k}\}\), with \(128\mathrm{k}\) as the upper cap. This policy avoids assigning an artificially restrictive context window while preserving meaningful context pressure.

\paragraph{Planner inference and output handling.}
All planner models receive the same serialized problem representation and the same planning instructions, without model-specific prompt tuning. We use a decoding temperature of \(0.1\) and a maximum completion length of \(16{,}000\) tokens. Each model--problem pair receives one primary planning call. If the response cannot be parsed or fails plan validation, the model receives one repair call containing the validation errors and its previous response. We do not manually edit or complete model-generated workflows. Plans that remain invalid after this repair attempt are treated as invalid according to the rule in Appendix~B. Agent identifiers are deterministically compacted after script expansion, so unused agent slots do not contribute to the reported active-agent count or simulation cost.

\paragraph{Compression and cost accounting.}
For the robust, balanced, and fragile compression classes, retaining a fraction \(e\) of a package preserves quality according to \(e^\gamma\), with
\(\gamma\in\{0.35,0.85,1.25\}\), respectively. Their minimum admissible retention ratios are \(0.08\), \(0.22\), and \(0.55\). When an agent exceeds \(L\), the simulator repeatedly compresses eligible memory items, prioritizing large and compression-robust packages while protecting fragile information. Each compression event adds token cost equal to \(2.5\%\) of the package size before that event and one unit of simulated time. A handoff of \(x\) retained tokens contributes \(x\) communication tokens and
\(\max(1,\lceil x/5000\rceil)\) units of time. Each active agent additionally incurs \(1{,}200\) startup tokens and two units of startup time.

\paragraph{Missing transfers and controlled sweeps.}
Unless otherwise specified, an omitted cross-agent dependency is modeled as a quality-only missing transfer with \(\lambda=0.5\): the upstream contribution is discounted by this factor without adding communication tokens, execution time, or stored context. This isolates information loss from additional resource penalties. Shared LLM- and tool-rate-limit waiting is disabled in the reported isolated-orchestration experiments, although the corresponding metadata are retained for optional contention studies. The agent-cap sweep uses
\(A_{\max}\in\{1,2,4,8,16,32,64\}\), while the context sweep uses
\(L\in\{16\mathrm{k},32\mathrm{k},64\mathrm{k},128\mathrm{k}\}\).
Both sweeps reuse the same 50 problems, sampled once with seed \(20260710\) and allocated as evenly as possible across the four nominal task scales. The extreme-scale study uses ten fixed problems at each of \(n\in\{200,500,1000\}\) and \(A_{\max}=100\). Since simulation is deterministic, no repeated simulator runs are required for a fixed problem and plan.

\section{H. Prompt Example}
\label{sec:prompts}

\subsection{I. DAG Generation Prompt}
This subsection provides the illustrative examples of the prompts adopted used in \bench{}. 
At runtime, the concrete seed record, current DAG, judge feedback, target task count, target parallelism bucket, and schema fields are inserted into these templates.

\subsubsection{Seed Filtering Prompt}
\label{app:prompt-seed-filtering}

\begin{promptbox}{Seed filtering prompt}
Decide whether the following raw dataset sample is suitable as a seed
question for an orchestration benchmark.

We will expand it into DAGs with 10/20/50/100 subtasks. A suitable seed
should:
1. have a clear final deliverable or final answer;
2. naturally decompose into a multi-stage workflow;
3. require at least two action types, such as evidence collection,
   code or data processing, writing, verification, or review;
4. have dependencies expressible as a DAG;
5. not mainly depend on private content, inaccessible environments, or an
   empty/overly short prompt.

Output only JSON:
{
  "accepted": true,
  "reason": "short reason",
  "suggested_workflow": ["stage 1", "stage 2", "stage 3"],
  "risk": "low | medium | high"
}
\end{promptbox}

\subsubsection{Refinement Prompt}
\label{app:prompt-refinement}

\begin{promptbox}{Refinement prompt}
Continue refining the following task DAG.

Rules:
1. Expand nodes that are too coarse, information-dense, or important for
   parallel scheduling. Multiple nodes may be expanded in one round.
2. The revised graph must remain an acyclic DAG with exactly one terminal
   delivery task.
3. Stay faithful to the original seed. Do not drift into another task.
4. Do not copy hidden answers, expected outputs, SQL, rubrics, or numeric
   answers from the source dataset into task descriptions.
5. Avoid duplicate, meaningless, or purely count-padding subtasks.
6. Hit the round target exactly when possible. Never exceed the final target.
   If the final target has already been reached, keep the node count fixed
   and only repair dependencies, task boundaries, stages, and metadata.
7. Preserve and update input tokens, execution tokens, result tokens,
   time cost, and compression class.
8. If a target parallelism bucket is specified, move k99 toward the target
   range without merely increasing layer width.
9. Judge feedback is an acceptance signal and targeted guidance, not a planner.
10. When expanding parallel work packages, preserve branch structure
   and avoid serializing independent work packages.
11. Output the complete revised JSON graph. Do not output a patch or markdown.
\end{promptbox}

\subsubsection{Semantic Judge Prompt}
\label{app:prompt-semantic-judge}

\begin{promptbox}{Semantic judge prompt}
Review whether the following task DAG is suitable as an orchestration
benchmark problem.

Check:
1. Do not repeat mechanical structural checks, but reject if the structural
   verifier reports hard issues.
2. Each subtask must be executable, not a placeholder.
3. Flag obvious duplication, over-splitting, under-splitting, or meaningless
   subtasks.
4. Check whether dependency edges are semantically natural and whether edges
   are missing or wrongly connected.
5. Check whether compression classes match task semantics.
6. Check whether token counts and time costs are plausible.
7. If the graph has not reached the target count but is structurally and
   semantically sound, it may be accepted with targeted refinement suggestions.
8. If the target count is reached and k99 is outside the requested bucket,
   reject and suggest dependency or boundary repairs.
9. High parallelism is part of the requested benchmark specification; do not
   reject a wide DAG merely because it is wide.

Output only JSON:
{
  "accepted": true,
  "issues": [],
  "warnings": [],
  "redundant_tasks": [],
  "suggested_refinements": []
}
\end{promptbox}

\subsubsection{Planner Evaluation Prompt}
\label{app:prompt-planner}

\begin{promptbox}{Planner evaluation prompt}
You are a dynamic workflow orchestrator. Output only JSON, no markdown and
no explanation. Write a small workflow_script rather than listing every
task assignment. The evaluator will deterministically expand the script
into a concrete assignment plan. The interpreter will not auto-complete
cross-agent transfers and will not choose compression ratios for you.

You receive the full DAG. deps are required upstream tasks. compression is
the sensitivity of the upstream result when passed to downstream tasks.
You also receive context_limit and max_agents.

Script semantics:
1. agent_pools define named pools; total count must not exceed max_agents.
2. rules are matched in order; assigned tasks are not overwritten.
3. match may use stage, stages, stage_regex, stage_contains, task_ids, or all.
4. strategy may be round_robin, dependency_grouping, load_balance, or single.
5. If parent and child are assigned to different agents, you must add explicit
   transfers or transfer_rules. Missing transfers are recorded and penalized
   through quality loss.
6. transfer_rules expand over dependency edges. cross_agent_only=true means
   transfer only when parent and child are on different agents.
7. compression must be explicitly set. Higher compression ratios preserve
   quality but cost more communication tokens. Choose the ratios yourself.
8. Use stage or regex rules to express a dynamic workflow; still cover the
   key cross-agent dependency transfers.

Output JSON with agent_pools, rules, default, and transfer_rules.
\end{promptbox}

\subsection{II. DAG Validation Prompt}
The LLM judge evaluates semantic validity that is difficult to capture with deterministic rules. 
The judge does not receive the full JSON for large DAGs. 
Instead, we compact the DAG into problem metadata, graph statistics, stage counts, largest layers, a rule-based validation snapshot, and representative task rows. 
Small DAGs are shown nearly in full. 
For large DAGs, task rows are sampled from the topology head, topology tail, widest layer, stage representatives, and stride samples. 
The hidden row count is included only for prompt-budget accounting and is not treated as missing subtasks.

\begin{promptbox}{LLM-as-judge prompt}
System:
You are a rigorous DAG-validity judge for benchmark orchestration. Output JSON only.

User:
Evaluate whether this DAG is a reasonable orchestration decomposition for the benchmark task.

Judge from a reviewer perspective. Focus on:
1. coverage: the DAG covers the original goal at an appropriate level of detail.
2. dependency_correctness: prerequisite edges and stage order are plausible.
3. redundancy: the DAG is not filled with duplicate or near-duplicate subtasks.
4. missing_subtasks: no obvious core stages are absent.
5. parallelism_plausibility: parallel branches are meaningful and have plausible integration.

All numeric scores are quality scores where 100 is best and 0 is worst.
For missing_subtasks, 100 means no obvious missing core stage/subtask type; 0 means severe missing core workflow stages.
Do not invert any rule percentage: hard-rule percentages are also quality percentages where higher is better.

Do not evaluate whether the final answer is correct, and do not evaluate time/token prediction ability.
For large DAGs, the task list is compacted; use stage/layer summaries and the sampled tasks to judge global validity.
The payload may contain unshown_task_row_count _for_prompt_budget_only. This is only the number of task rows hidden to fit the prompt.
It is not a count of missing DAG subtasks and must not reduce missing_subtasks by itself.
The hard-rule snapshot is diagnostic evidence, not an answer key.

Return only a JSON object with this schema:
{
  "accepted": "boolean; true only if the DAG is reasonable enough for benchmark use",
  "scores": {
    "coverage": "integer 0-100",
    "dependency_correctness": "integer 0-100",
    "redundancy": "integer 0-100",
    "missing_subtasks": "integer 0-100",
    "parallelism_plausibility": "integer 0-100",
    "overall": "integer 0-100"
  },
  "issues": "list of short strings; concrete problems, not generic caveats",
  "strengths": "list of short strings",
  "confidence": "low|medium|high",
  "rationale": "one concise paragraph"
}

DAG payload:
{... compacted DAG ...}
\end{promptbox}


\end{document}